\documentclass[10pt,twocolumn,letterpaper]{article}

\usepackage{wacv}
\usepackage{times}
\usepackage{epsfig}
\usepackage{graphicx}
\usepackage{amsmath}
\usepackage{amssymb}
\usepackage{booktabs}
\usepackage[sort,square,numbers]{natbib}
\usepackage[export]{adjustbox}
\usepackage{float}
\usepackage[title]{appendix}

\usepackage{listings}
\usepackage{xcolor}

\definecolor{codegreen}{rgb}{0,0.6,0}
\definecolor{codegray}{rgb}{0.5,0.5,0.5}
\definecolor{codepurple}{rgb}{0.58,0,0.82}
\definecolor{backcolour}{rgb}{0.95,0.95,0.92}

\lstdefinestyle{mystyle}{
    commentstyle=\color{codegreen},
    keywordstyle=\color{magenta},
    numberstyle=\tiny\color{codegray},
    stringstyle=\color{codepurple},
    basicstyle=\ttfamily\footnotesize,
    breakatwhitespace=false,         
    breaklines=true,                 
    captionpos=b,                    
    keepspaces=true,                 
    numbers=left,                    
    numbersep=5pt,                  
    showspaces=false,                
    showstringspaces=false,
    showtabs=false,                  
    tabsize=2
}

\def\wacvPaperID{2086} 

\wacvalgorithmstrack   

\wacvfinalcopy 

\ifwacvfinal
\usepackage[breaklinks=true,bookmarks=false]{hyperref}
\else
\usepackage[pagebackref=true,breaklinks=true,colorlinks,bookmarks=false]{hyperref}
\fi

\begin{document}

\title{
OCR-VQGAN: Taming Text-within-Image Generation
}

\author{Juan A. Rodriguez$^1$, David Vazquez$^2$, Issam Laradji$^2$, Marco Pedersoli$^3$, Pau Rodriguez$^2$\\
$^1$Computer Vision Center, Barcelona, $^2$ServiceNow Research, $^3$ÉTS Montréal\\
{\tt\small joanrg.ai@gmail.com}
}
\maketitle
\thispagestyle{empty}

\begin{abstract}
    
   Synthetic image generation has recently experienced significant improvements in domains such as natural image or art generation. However, the problem of figure and diagram generation remains unexplored. A challenging aspect of generating figures and diagrams is effectively rendering readable texts within the images. To alleviate this problem, we present OCR-VQGAN, an image encoder, and decoder that leverages OCR pre-trained features to optimize a text perceptual loss, encouraging the architecture to preserve high-fidelity text and diagram structure. To explore our approach, we introduce the Paper2Fig100k dataset, with over 100k images of figures and texts from research papers. The figures show architecture diagrams and methodologies of articles available at arXiv.org from fields like artificial intelligence and computer vision. Figures usually include text and discrete objects, e.g., boxes in a diagram, with lines and arrows that connect them. We demonstrate the effectiveness of OCR-VQGAN by conducting several experiments on the task of figure reconstruction. Additionally, we explore the qualitative and quantitative impact of weighting different perceptual metrics in the overall loss function. We release code, models, and dataset at \textcolor{magenta}{\url{https://github.com/joanrod/ocr-vqgan}}.
   
\end{abstract}

\section{Introduction}
Image synthesis efforts in the recent literature have achieved impressive results in the domain of natural images. Some examples are face generation, landscapes, and art~\citep{StyleGan2, DALLE2, Imagen, Parti, CogView, LatentDiffusion}. Current methods can generate high-resolution and realistic images, allowing creators to guide the generation process using text descriptions and other conditioning modalities~\citep{make_a_scene}. Within the next few years, text-to-image generation models are set to enhance and complement creative processes in many fields including art, design, video games, and content creation.
\begin{table}[t]
    \begin{center}
    \setlength{\tabcolsep}{2pt}
    \footnotesize
            \begin{tabular}{c|c|c}
                \toprule
               \textbf{Ground truth} & \textbf{VQGAN} & \textbf{OCR-VQGAN} (ours)\\
                \midrule
                
                \includegraphics[width=0.30\linewidth]{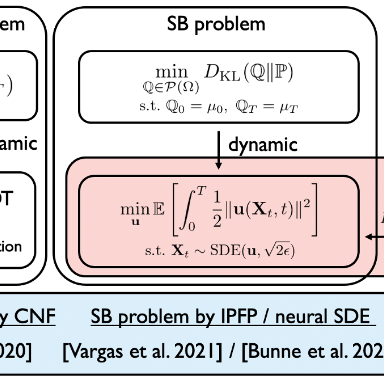} &
                \includegraphics[width=0.30\linewidth]{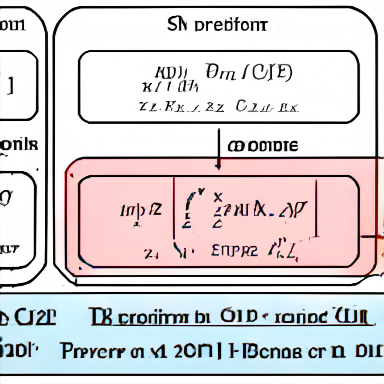} &
                \includegraphics[width=0.30\linewidth]{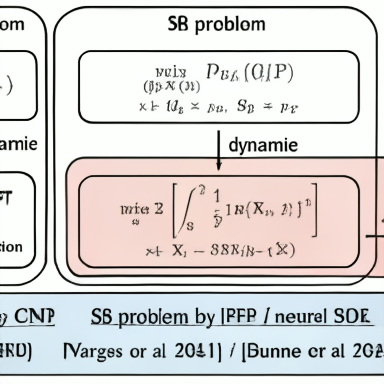} \\ \midrule
                
                \includegraphics[width=0.30\linewidth]{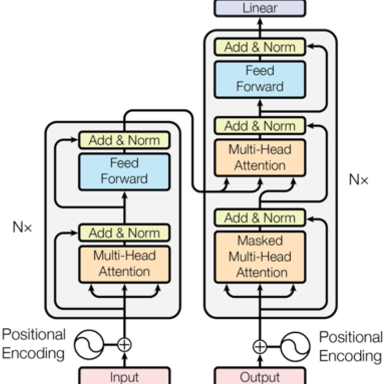} &
                \includegraphics[width=0.30\linewidth]{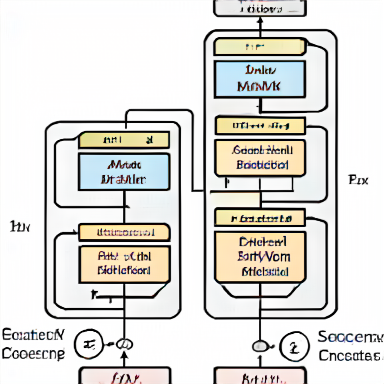} &
                \includegraphics[width=0.30\linewidth]{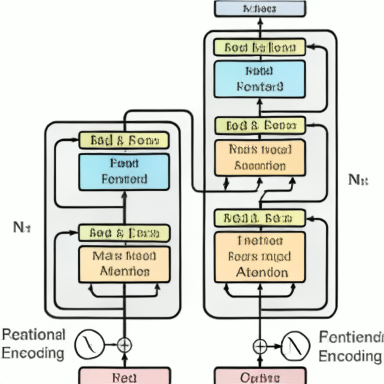} \\ \midrule
                
                \includegraphics[width=0.30\linewidth]{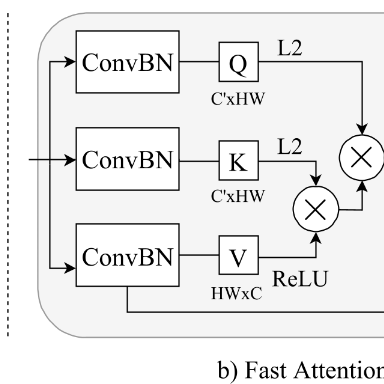} &
                \includegraphics[width=0.30\linewidth]{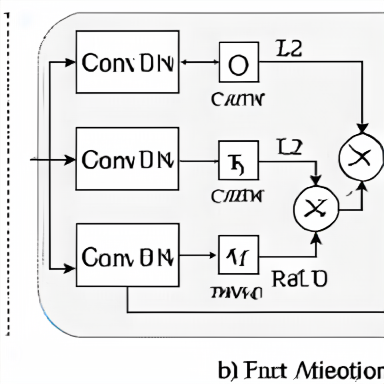} &
                \includegraphics[width=0.30\linewidth]{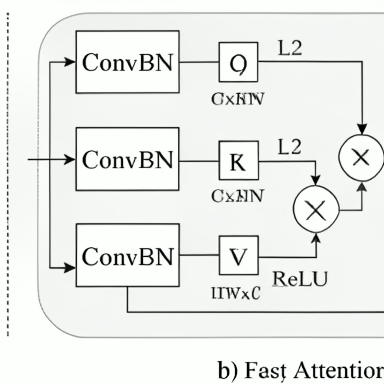} \\
                
                \bottomrule
            \end{tabular}
    \end{center}
    \caption{ Qualitative comparison for the task of figure reconstruction. OCR-VQGAN outperforms VQGAN at capturing text and symbol details.}
    \label{table:comparison_fig1}
\end{table}

However, a common deficiency of current methods like Parti~\citep{Parti} or Imagen~\citep{Imagen} is that they tend to fail at text rendering within images, as highlighted in~\citep{Parti} (limitations section). Current image generation systems trained on natural images do not produce the desired results for applications like structured diagram generation or automatic slide edition. This shortcoming makes recent text-to-image generation methods unsuitable to aid people to design and create figures and diagrams for their work. Thus, in this work, we tackle the problem of generating text within images.
We focus on the generation of paper figures, which typically contain a visually summarized description of methods or architectures presented in academic papers. The use of figures in research publications has become widespread in many fields (we focus on Deep Learning (DL) and Computer Vision (CV)), allowing researchers to describe ideas compactly. People spend a significant amount of time building appealing and understandable figures. Therefore, tools for assisting this process will be beneficial. 

Although figures present a more precise structure and composition than natural images (i.e., connectivity between objects, textual descriptions, legends), \citet{Diag2graph} highlight that there is little agreement on how these figures should be created, as they do not follow a specific set of rules for formatting or structuring information (see 
Figure~\ref{fig:Paper2Fig100k}). Writers must define many parameters when generating figures, such as size, the position of shapes, colors, connections, or text styles. Hence, building image generation models in the space of figures is rather time-consuming, as there is significant variability. In this work, we tackle the problem of generating figures by (i) introducing a novel dataset of figure images and text pairs from research papers and (ii) addressing the text rendering problem of current image encoders.

We acquire a large database of figure images and texts from publicly available research papers on arXiv.org, which we call Paper2Fig100k. The dataset contains over 100k pairs of images and text captions (see Section~\ref{sec:Paper2Fig100k}), designed for the task of text-conditional generation of figures, which are important components in
many real-life applications such as architectural blueprints, project designs, and research papers. The proposed dataset allows tackling the problem of text rendering in diagrams, which was not possible in previous scene-text datasets.

We construct a rich image encoder for figures. We build upon the VQGAN~\citep{VQGAN} approach, based on learning a codebook of patch embeddings, exploiting perceptual and patch-based adversarial objectives. Our method adds an OCR perceptual loss term to the training objective that minimizes the distance between original and reconstructed images in the feature space. To this end, we add a pre-trained OCR detection model, encouraging the learned embeddings to represent clear text and diagrams (see Table~\ref{table:comparison_fig1}). 

Since we focus on compressing images in the domain of figures, we need to impose the model to learn the most frequent patches and local patterns present in the dataset, building a rich discrete latent representation (i.e., a codebook of patches). For instance, our encoder needs to represent a variety of text possibilities, such as different sizes, styles, fonts, and the orientation of characters. Text and background colors also need to be considered, as well as the sharpness of arrows, lines, or geometric shapes.

Traditional image reconstruction metrics heavily rely on L2 similarities in the pixel space, which tend to fail at quantifying high-level perceptual similarity. For example, a simple shift of the image in the horizontal axis makes a point-wise L2 distance give incorrect results. In contrast, humans can detect patterns and the overall structure of images, allowing them to make a better perceptual assessment. This can be achieved by minimizing a perceptual loss in the feature space, using models pre-trained for image recognition.

Our contributions can be summarized as follows. We propose:
\begin{itemize}
\item Paper2Fig100k, a novel dataset for the task of text-to-image generation, composed of texts and images acquired from publicly available research papers;
 \item OCR-VQGAN, an image encoder focused on synthesizing images of figures, preserving text-within-images and diagram structure; and
 \item an OCR perceptual loss and OCR similarity metric (OCR-SIM), devoted to measuring the perceptual distance of images with respect to an OCR pre-trained model.
\end{itemize}

The rest of the paper is structured as follows. Section 2 gives an overview of the recent literature on deep generative models, image encoders, and diagram-based tasks and datasets. Section 3 describes Paper2Fig100k, a novel dataset of research figures and texts. In Section 4 we propose OCR-VQGAN, an image encoder focused in preserving textual clarity within images. Experiments are conducted in Section 5. In Section 6 conclusions are drawn. Finally, in the final section, we reflect on the ethical and social impact of text-to-image generation.

\section{Related Work}
\paragraph{Text-to-image synthesis.}
Recent text-to-image generation methods have achieved outstanding image generation quality even for unseen concept compositions. Works like DALLE~\citep{DALLE}, DALLE-mini~\citep{DALLE_Mini}, or Parti~\citep{Parti} pose the task as a language modeling problem, using Transformer~\citep{vaswani} architectures to learn the relationships between text and visual information. This family of methods relies on image encoders (or tokenizers) such as VQ-VAE~\citep{VQVAE} or VQGAN~\citep{VQGAN}, that convert images into a sequence of tokens, allowing to treat both text and image modalities as a sequence-to-sequence task. Another family of methods that have gained popularity use diffusion methods~\citep{Diffusion} to directly generate images from text embeddings. Works like DALLE2~\citep{DALLE2} and Imagen~\citep{Imagen}, use CLIP~\citep{CLIP} and~\citep{T5} text encoders to condition the diffusion process. Latent Diffusion~\citep{LatentDiffusion} incorporates a BERT tokenizer and a more flexible cross-attention mechanism to condition the diffusion process. A weakness of current text-to-image synthesis methods is that they struggle to generate text~\citep{Imagen,Parti}. In this work, we improve the image encoder module to support text generation within images.

\paragraph{Image tokenizers.}
Vector Quantized Variational Autoencoder (VQVAE)~\citep{VQVAE,razavi2019generating} is a popular approach for learning discrete representations of images. VQ-based methods learn a codebook of discrete latent embeddings and use a nearest neighbor algorithm to map continuous latent features to discrete embeddings. They propose to model the data distribution by means of autoregressive density estimation, using causal convolutional kernels.
VQGAN uses the quantization procedure of VQVAE and improves the richness of VQVAE's learned codebook. The authors of VQGAN modify the training objective using a VGG perceptual loss~\citep{LPIPS} and a patch-based adversarial module to obtain high-quality embeddings. They demonstrate how learning a rich codebook of image patches is crucial in order to perform high-resolution image synthesis. Although VQGAN improves the reconstruction quality of VQVAE, the model still struggles to draw text (see Table~\ref{table:method_comparison}. In this work, we include an additional perceptual loss that improves text reconstruction.

\paragraph{Perceptual-based reconstruction.}
Perceptual similarity losses are common in the field of Style Transfer~\citep{StyleGAN,Perceptual_Johnson,larsen}. \citet{LPIPS} prove the effectiveness of using learned perceptual losses as a similarity objective, which is evaluated in the feature space. Their work shows how VGG16~\citep{vgg} pre-trained on Imagenet~\citep{imagenet} can be used as reconstruction loss, capturing differences in perceptual similarity. Perceptual losses pre-trained on Imagenet are adequate for measuring differences between natural images. However, they are inadequate to measure distances in text generation, since text recognition is not an objective of the Imagenet. Here we alleviate this problem by introducing an additional perceptual loss obtained from a model trained for text detection Optical Character Recognition (OCR)~\citep{CRAFT_OCR}.

\paragraph{Related datasets and tasks.}
Some works have been proposed in the domain of document and figure analysis, mainly focused on classification~\citep{DocFigure} and object detection~\citep{Diag2graph} tasks or visual question answering~\citep{FigureQA}. Most of the available datasets contain many types of figures such as tables, flow charts, and different types of plots. \citet{SciCap} introduces a dataset of figures and texts for the task of image captioning, which contains $60,000$ samples of all available figures in scientific papers (e.g., scatter plots, bar plots, flowcharts, equations). \citet{FigCap} also approaches figure captioning by introducing the FigCAP dataset, containing samples from many types of figures. To the best of our knowledge, there are no publicly available datasets focused on diagram figures. We propose the construction of a new dataset of figure diagrams and texts from research articles.

\section{Paper2Fig100k dataset}\label{sec:Paper2Fig100k}
\begin{figure*}[t]
\begin{center}
\includegraphics[width=1.0\linewidth]{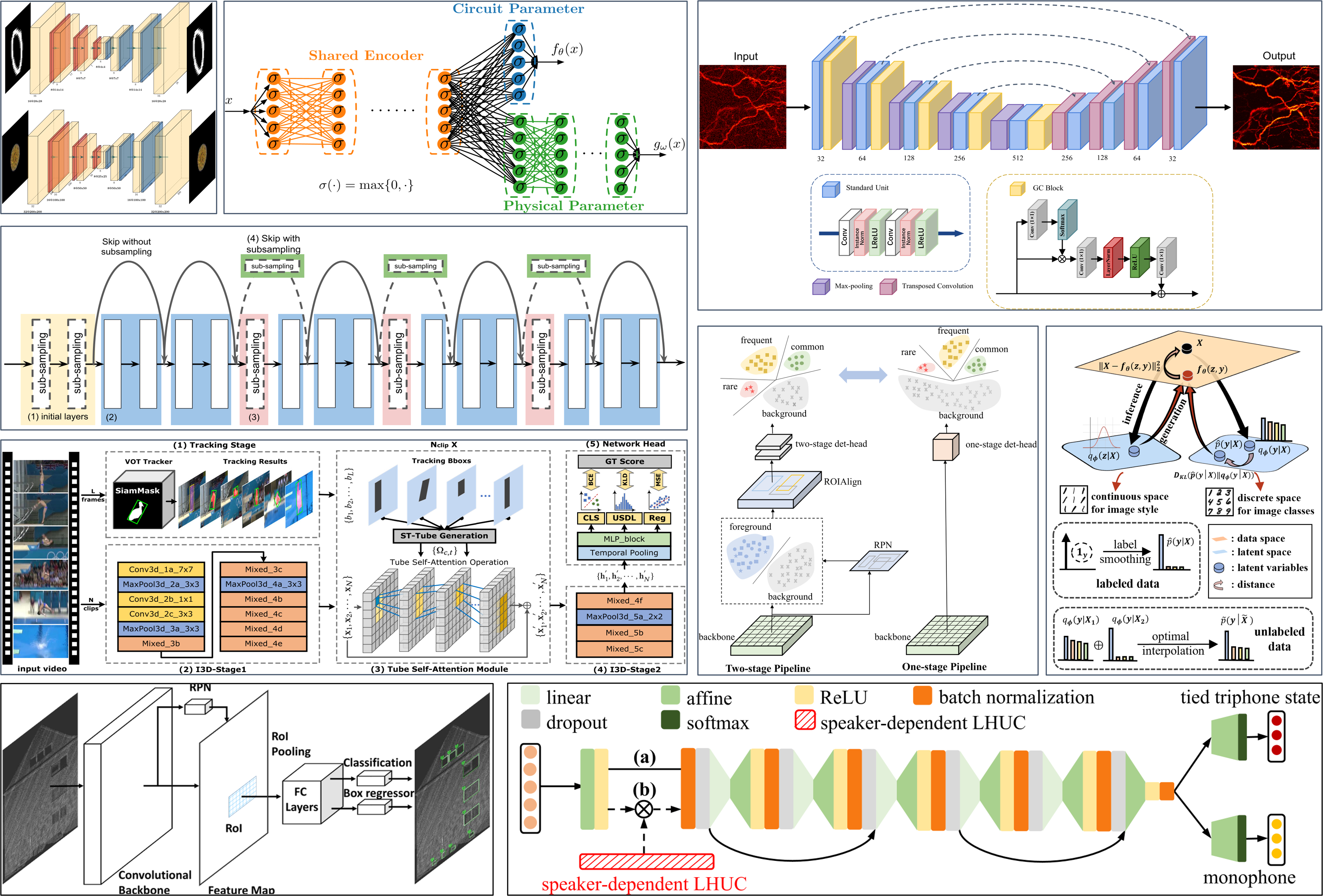}
\end{center}
  \caption{Samples from Paper2Fig100k dataset. We show how the samples have high variability in aspect ratio, image resolution, text and diagram sizes, or amount of information displayed. Unlike natural images, figures contain fine-grained information details that are relevant for a complete understanding of the diagram.}
\label{fig:Paper2Fig100k}
\end{figure*}

In order to accomplish text-to-figure generation, and to address the lack of publicly available datasets for the task, we present the Paper2Fig100k dataset. It consists on $102,453$ pairs of images and texts from $69,413$ papers. The data is split into a training set of $81,194$ samples and a test set of $21,259$ samples. While the proposed dataset is meant for text-to-figure generation, it can also be used to train the first stage of a text-to-image pipeline (image encoder). Paper2Fig100k can also be used for image-to-text generation (reverse process) and multi-modal vision-language tasks. Samples from the dataset are shown in Figure~\ref{fig:Paper2Fig100k}.

\begin{figure}[t]
\begin{center}
\includegraphics[width=\linewidth, height=0.8\linewidth, trim={0 0 0 0},clip]{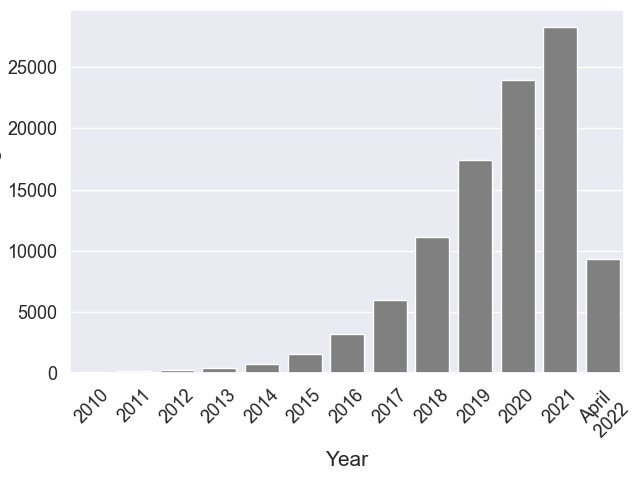}
\end{center}
  \caption{In this plot we show the number of figures extracted per year, ending in April 2022. This figure shows the exponential increase in research publications.}
\label{fig:year-fig}
\end{figure}

Paper2Fig100k contains images of architectures, diagrams, and pipelines (generally referred to as figures), with detailed text captions acquired from public research papers at arXiv.org. It also includes OCR-detected bounding boxes and text transcriptions of figures, that can be used for hand-crafted attention and fine-grained text conditioning. As shown in Figure~\ref{fig:year-fig}, the dataset is expected to increase exponentially over the years.

\begingroup
\begin{table}[t]
\renewcommand{\arraystretch}{1.5}
    \begin{center}
            \begin{tabular}{l p{5cm}}
                \toprule
                \multicolumn{2}{c}{Input image} \\
                \multicolumn{2}{c}{\includegraphics[width=0.9\linewidth]{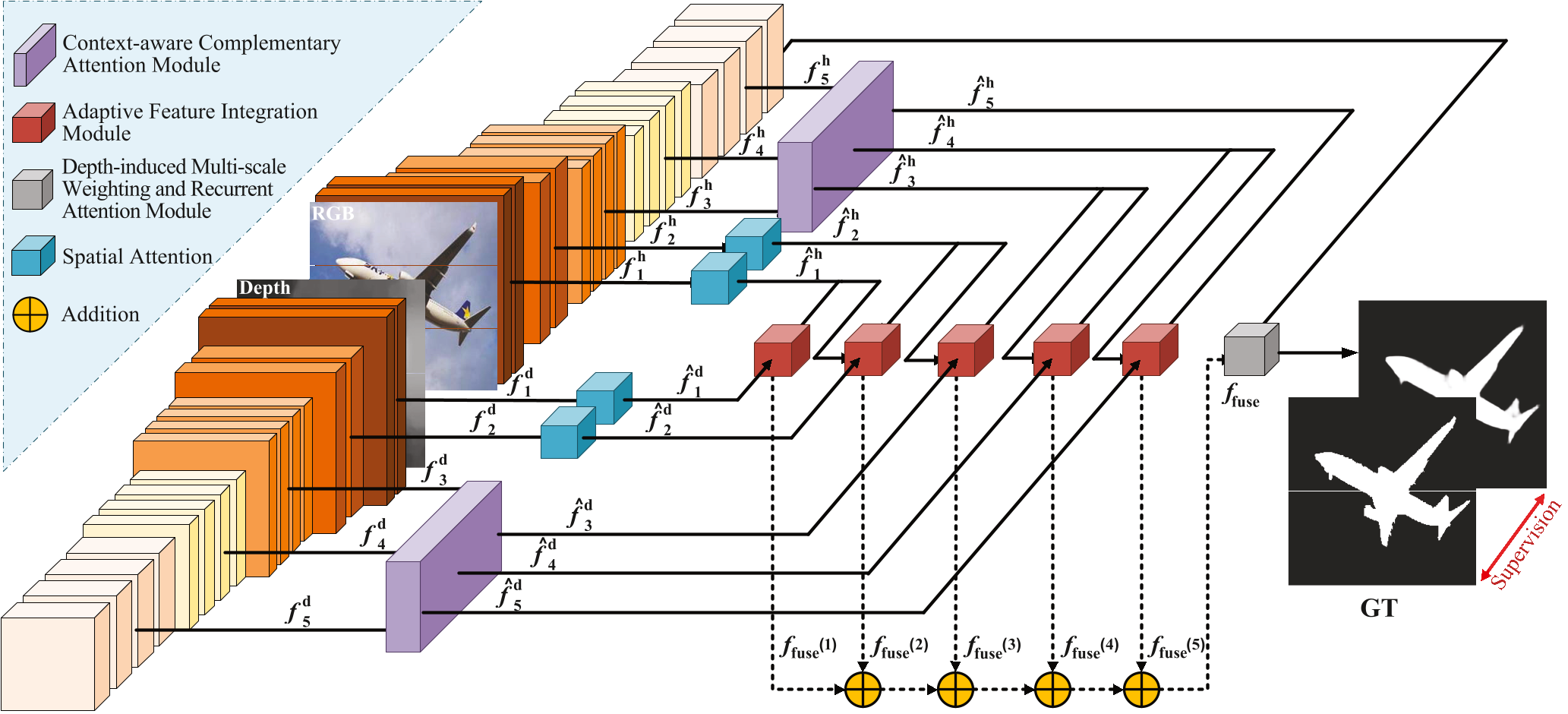}}\\
                \midrule
                Prompt Modality & Examples \\
                \midrule
                Caption & \textit{Figure 2: An overview of our network. We propose a Complementary Attention and Adaptive Integration Network}\\
                
                References & \textit{Fig. 2 shows an overview of CAAI-Net, which is based on a two-stream structure for RGB images and depth maps. As can be observed, (...).}\\
                OCR keywords & \textit{Context-aware Complementary Attention Module, (...).}\\
                \bottomrule
            \end{tabular}
    \end{center}
    \caption{
        Example of our captioning system for the task of text-conditional image synthesis. \textbf{Note} that, the figure shown at the top of the table is a sample from the dataset \protect\footnotemark.
    }
    \label{table:Paper2Fig100k_captions}
\end{table}
\endgroup
\setcounter{footnote}{1} 

\footnotetext{The example figure was extracted from the paper \textit{
                Towards Accurate RGB-D Saliency Detection with Complementary Attention and Adaptive Integration
                }}

\subsection{Data acquisition pipeline} 
The dataset was acquired using the API and metadata offered by \textit{arXiv dataset}~\citep{arxiv_kaggle}, which includes both paper metadata (e.g., title, abstract, authors, or research fields) and tools for downloading the papers in pdf format via Google Cloud. \textit{arXiv dataset} is updated weekly, offering more than $1.7$ million papers across all STEM fields.

The complete set of papers available covers a vast amount of fields in the arXiv taxonomy\footnote{arXiv follows a standardized taxonomy to encode  field categories, \url{https://arxiv.org/category_taxonomy}}, therefore we filter and keep papers in the categories of Machine Learning (cs.LG), Artificial Intelligence (cs.AI), Computer Vision and Pattern Recognition (cs.CV), and Computation and Language (cs.CL). We downloaded all papers published after January 2010, which represent a total of $183,427$ papers.

Papers in pdf format are processed and parsed using the GROBID~\citep{GROBID} open-source library, which allows to extract and organize texts and images of pdf files, mostly focused on technical or scientific documents. It is based on a cascade of models for object detection such as conditional random fields (CRF). The software is production ready and very competitive in terms of processing speed (it is capable of processing $10.6$ pdf per second). We release the pipeline at \textcolor{magenta}{\url{https://github.com/joanrod/paper2figure-dataset}}.

\subsection{Heuristics for obtaining figures}
The $183,427$ downloaded papers contain $\sim1.6M$ images. However, many of them contain qualitative results or other kinds of natural images that we want to avoid. Since we are interested in figure diagrams, we apply simple heuristics to keep only figures describing architectures or methods, and remove figures related to results or examples. The result is a set of 102,453 images.

We use text-based heuristics using figure captions. We keep figures containing strings in the caption such as ``architecture", ``model diagram'' or ``pipeline". We remove figures with words like ``table", ``results" or ``example", which may not correspond with the desired figures. Gray-scale histograms are used to remove outliers such as blank (all white) or natural images (almost no white). 

We process images with an OCR detection and recognition system (EasyOCR), based on the  CRAFT~\citep{CRAFT_OCR} OCR detector and the CRNN~\citep{EasyOCR_CRNN} text recognizer. Note that this process is applied once and it is independent of the OCR-VQGAN method later introduced in this work. The goal is to obtain textual tags and captions to automatically annotate samples and use them for text conditioning.

\subsection{Prompt modalities}
Text captions generally encode the same information that is represented in the figure, hence text-conditional image synthesis can be achieved. There are many options for conditioning figure generation based on the text of a paper. As figures usually describe the methods presented in their papers, we propose pairing figure images with the text in \textit{methodology} sections. We also explore other conditioning information such as figure captions and keywords extracted from an OCR model.

Concretely, we define three prompting modalities for future research in text-to-image generation: caption, references, and OCR keywords. Captions are obtained directly from the figure caption. References are extracted from paragraphs in the paper referencing the figure. OCR keywords are a concatenation of the texts detected by the OCR model. Table \ref{table:Paper2Fig100k_captions} exemplifies the kind of captions in the dataset.

\section{Method}
Our goal is to train a figure-based image encoder (tokenizer) capable of transforming images into sequences of discrete tokens, and an image decoder (detokenizer) that reconstructs figures from tokens, preserving details of text and diagram structure. Tokens are indices of a learned codebook of patch embeddings in the discrete latent space, that encodes the patches in the original image. When encoding an image, individual patch embeddings are assigned to the nearest codebook entry. To solve the reconstruction task at hand, the method needs to learn the most relevant and realistic patches within our dataset, and assign discrete embeddings in the codebook. To this end, the proposed OCR-VQGAN encoding and decoding pipeline uses a patch-based adversarial procedure, a VGG-based perceptual (LPIPS) loss, and a novel OCR perceptual loss.

\subsection{OCR-VQGAN}
We leverage VQGAN's image encoder~\citep{VQGAN} to learn a mapping from the image space to a discrete latent representation of tokens. To this end, the VQGAN architecture is composed of an image encoder, a decoder, and a vector quantization stage. The encoder is devoted to downsample an image $x \in \mathbb{R}^{H\times W\times 3}$ into discrete codes $z_q \in \mathbb{R}^{h\times w\times n_z}$, where $n_z$ is the size of the embedding space. One can simply describe each code with its codebook index and rearrange the discrete representation by a grid of shape $h \times w$. 

Using the same architecture and notation as~\citet{VQGAN}, the image encoder $E$ and decoder $G$ are convolutional neural networks aimed at learning a discrete codebook $\mathcal{Z} = \{z_k\}_{k=1}^{K} \subset \mathbb{R}^{n_z}$. In a forward pass, an image $x$ is approximated as $\hat{x}=G(q(E(x)))$, where $q$ is the quantization function, which performs a nearest neighbour operation. Instead of using an L2 loss, they use the LPIPS perceptual loss by~\citet{LPIPS}, which is better at capturing perceptually rich details of the image. Finally, VQGAN introduces a patch-based adversarial strategy with a discriminator $D$ that learns to distinguish real from fake patches, therefore the decoder $G$ gets better at generating realistic samples. 

The original VQGAN loss can be expressed as follows,
\begin{equation}
    \mathcal{L}_{\text{\textit{VQGAN}}} = \mathcal{L}_{\text{\textit{VQ}}}(E, G, \mathcal{Z}) + \lambda \mathcal{L}_{\text{\textit{GAN}}}(\{E, G, \mathcal{Z}\}, D),
\end{equation}
\label{eq:VQGAN_LOSS}
where $\mathcal{L}_{\text{\textit{VQ}}}$ is the vector quantization loss, and $\mathcal{L}_{\text{\textit{GAN}}}$ is a patch-based Hinge loss, and $\lambda$ is an adaptive weight for $\mathcal{L}_{\text{\textit{GAN}}}$ (refer to \citep{VQGAN} for detailed derivation).

\subsection{OCR Perceptual Similarity}
We propose a new OCR perceptual loss for rendering clear texts in generated images. We use a frozen pre-trained CRAFT~\citep{CRAFT_OCR} model, which is a text detector trained to localize individual characters in natural images. The model is based on VGG16~\citep{vgg} with batch normalization as backbone and uses a U-net~\citep{UNET} architecture with skip connections in the upsampling layers. The CRAFT model is kept frozen and adds 20M parameters to the VQGAN architecture, which is comparable to the 14M parameters of LPIPS.
\begin{figure}[t]
    \begin{center}
    \includegraphics[width=0.85\linewidth]{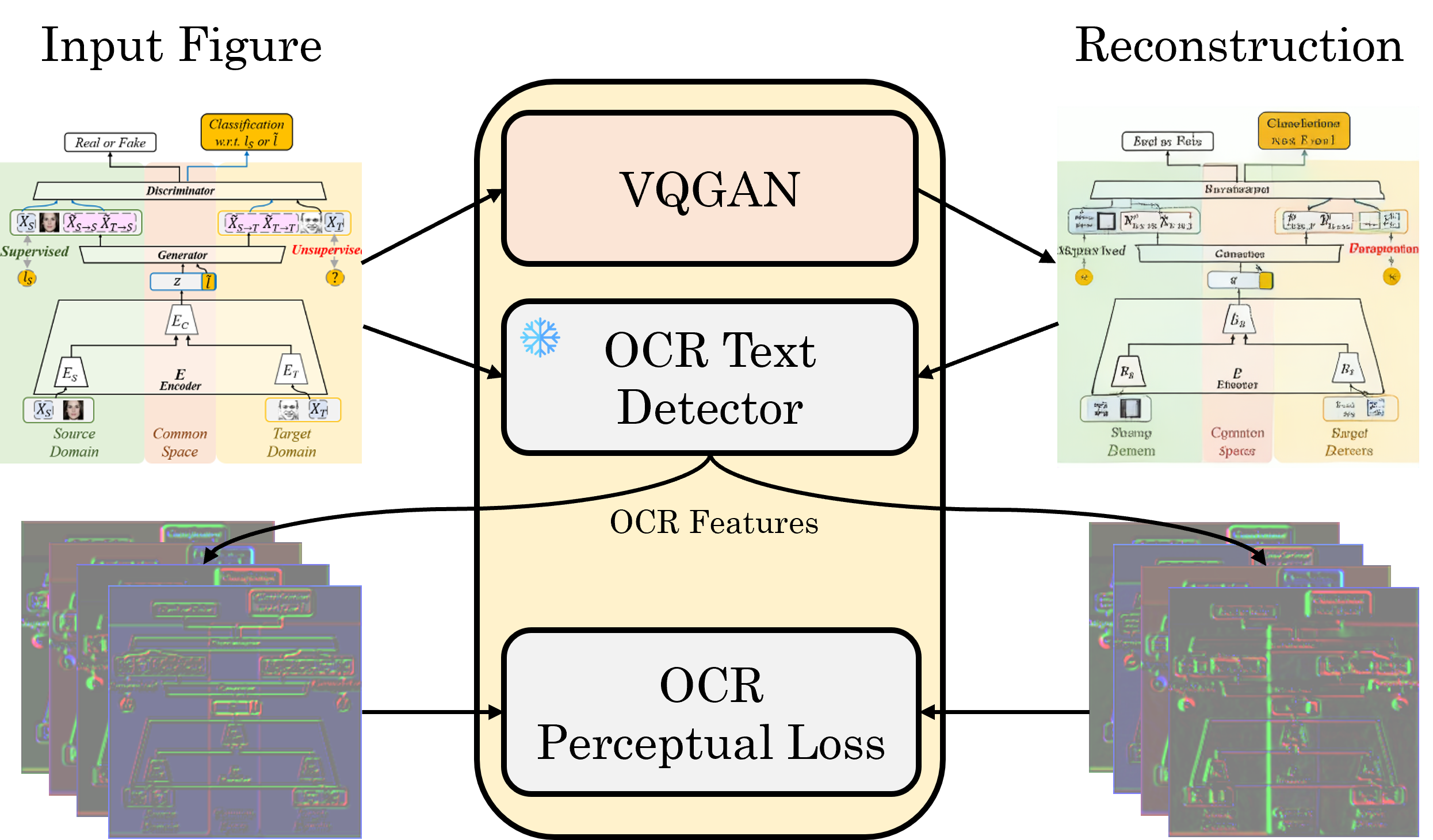}
    \end{center}
    \caption{OCR Perceptual loss computation in OCR-VQGAN. OCR feature maps are extracted at intermediate layers and the OCR loss is computed as depicted in equation~\ref{eq:OCR_LOSS}.}
    \label{fig:QualitativeComparisonMethods}
\end{figure}
As introduced by~\citep{LPIPS}, we forward input patches $x$ and reconstructed patches $x_0$, which represent the input and reconstructed images, through the OCR model, and extract $L$ feature maps from intermediate layers. Specifically, we store the activation map after each upsampling layer. Layers are then normalized in the channel dimension, denoted as $\hat{y}_{hw}, \hat{y}_{0hw} \in \mathbb{R}^{H_l\times W_l\times C_l}$ for each layer $l$. The OCR perceptual loss is expressed as
\begin{equation}
    \mathcal{L}_{ocr} = \sum_{l}\frac{1}{H_l W_l} \sum_{h, w} ||\hat{y}_{hw}^l - \hat{y}_{0hw}^l||_2^2.
\end{equation}
\label{eq:OCR_LOSS}
Instead of using a network $\mathcal{F}$ with weights $w_l$ to scale feature maps (proposed in LPIPS~\citep{LPIPS}), we simply average in the spatial dimension and sum over the channel dimension ($w_l=1 \forall l$). The OCR perceptual loss is added to the loss function in equation \ref{eq:VQGAN_LOSS}, which then defines the OCR-VQGAN loss. 
\begin{table}[t]
 \begin{center}
\begin{tabular}{lccc}
\toprule
Method & $\text{samples}/\text{s}$ &  dim $\mathcal{Z}$ & Params. \\
\midrule
VQVAE & 9.69 & 8192 & 97M\\
VQGAN & 8.17 & 16834 & 92M\\
OCR-VQGAN & 6.35 & 16834 & 112M\\
\bottomrule
\end{tabular}
\end{center}
\caption{Test time results using Paper2Fig100k test set. The test performs a forward pass of the samples and computes both LPIPS and OCR perceptual similarities, using only 1 V100 GPU.}
\label{table:overhead}
\end{table}

\begin{table}[t]
 \begin{center}
\begin{tabular}{lccc}
\toprule
Method & LPIPS$\downarrow$ & OCR-SIM$\downarrow$ & FID$\downarrow$ \\
\midrule
\multicolumn{4}{c}{\footnotesize\textbf{Paper2Fig100k}}\\ \midrule
VQVAE$_{\text{\textit{DALLE}}}$  & 0.10  & 0.87 & 9.91 \\
VQGAN$_{\text{\textit{Imagenet}}}$  & 0.12  & 1.04 &  6.68 \\ 
VQGAN$_{\text{\textit{Paper2Fig100k}}}$  & 0.15  & 1.18 & 4.37 \\ 
OCR-VQGAN  & \textbf{0.07} & \textbf{0.42} & \textbf{1.69} \\
\midrule
\multicolumn{4}{c}{\footnotesize\textbf{ICDAR13}}\\ \midrule
VQVAE$_{\text{\textit{DALLE}}}$  & 0.23  & 1.17 & 71.84 \\
VQGAN$_{\text{\textit{Imagenet}}}$  & 0.22  & 1.61 & 37.06 \\ 
VQGAN$_{\text{\textit{Paper2Fig100k}}}$  &  0.29  & 1.97 & 133.97 \\ 
OCR-VQGAN  & 0.36  & 1.26 & 84.77 \\
\bottomrule
\end{tabular}
\end{center}
\caption{Quantitative comparison of methods in the reconstruction task. The first part of the table corresponds to the method evaluation on Paper2Fig100k test set, and the second part corresponds to test results on the full ICDAR13.}
\label{table:results_p_losses}
\end{table}

\begin{table}[t]
 \begin{center}
 \resizebox{0.9\linewidth}{!}{
\begin{tabular}{ccccc}
\toprule
$w_{ocr}$ & $w_{vgg}$ & LPIPS$\downarrow$ & OCR-SIM$\downarrow$ & FID$\downarrow$ \\
\midrule
0.0 & 1.0  & 0.15  & 1.18 & 4.37 \\
\midrule
1.0 & 0.0  & 0.10  & 0.50 & 2.23 \\
\textbf{1.0} &\textbf{0.2} &\textbf{0.07} & \textbf{0.42} & \textbf{1.69}\\
1.0 & 0.5 & 0.08  & 0.46 & 1.84\\
1.0 & 0.8 & 0.09 & 0.50 & 2.03 \\
1.0 & 1.0 & 0.08 & 0.49 & 2.10 \\

\bottomrule
\end{tabular}
}
\end{center}

\caption{Results on Paper2Fig100k test set using OCR-VQGAN. We compare different weighting settings for VGG Perceptual loss ($w_{vgg}$) and  OCR Perceptual Loss ($w_{ocr}$).}
\label{table:results_method_comparison}
\end{table}
\begin{table*}[t]
    \begin{center}
    \setlength{\tabcolsep}{2pt}
        \resizebox{0.9\textwidth}{!}{
            \begin{tabular}{ccccc}
                \toprule
               \textbf{Ground truth} & \textbf{VQVAE$_{\text{\textit{DALLE}}}$} & \textbf{VQGAN$_{\text{\textit{Imagenet}}}$} & \textbf{VQGAN$_{\text{\textit{Paper2Fig100k}}}$} & \textbf{OCR-VQGAN} \\
                \midrule
                \multicolumn{5}{c}{\footnotesize\textbf{Evaluated on Paper2Fig100k}}\\
                
                \includegraphics[width=0.19\linewidth]{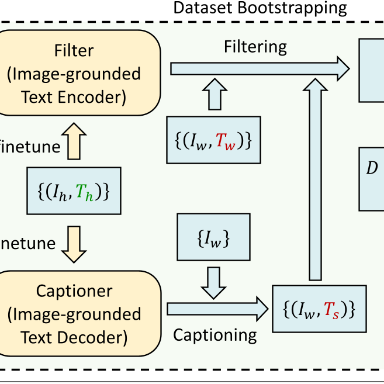} &
                \includegraphics[width=0.19\linewidth]{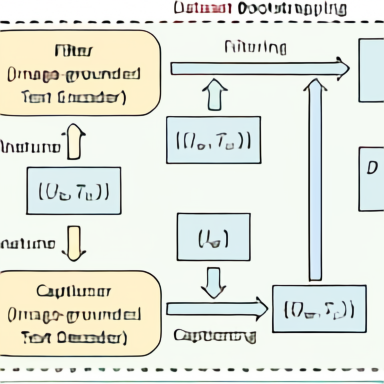} &
                \includegraphics[width=0.19\linewidth]{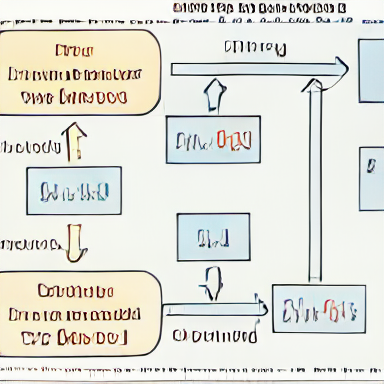} &
                \includegraphics[width=0.19\linewidth]{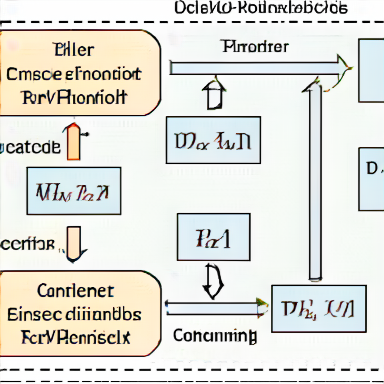} &
                \includegraphics[width=0.19\linewidth]{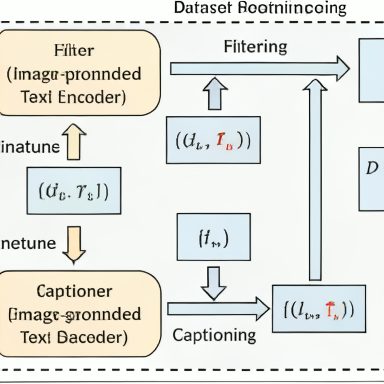}
                \\
                
                \includegraphics[width=0.19\linewidth]{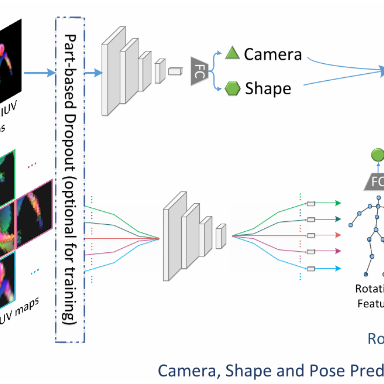} &
                \includegraphics[width=0.19\linewidth]{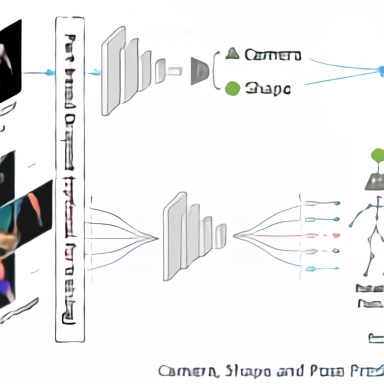} &
                \includegraphics[width=0.19\linewidth]{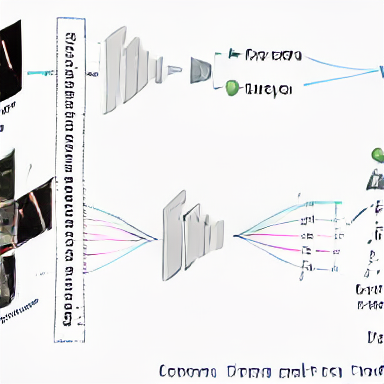} &
                \includegraphics[width=0.19\linewidth]{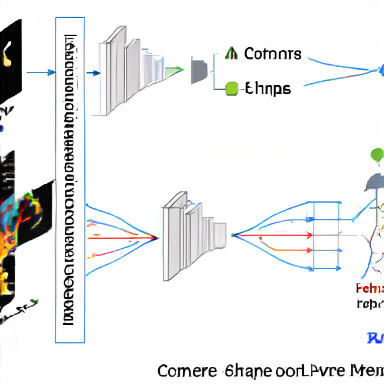} &
                \includegraphics[width=0.19\linewidth]{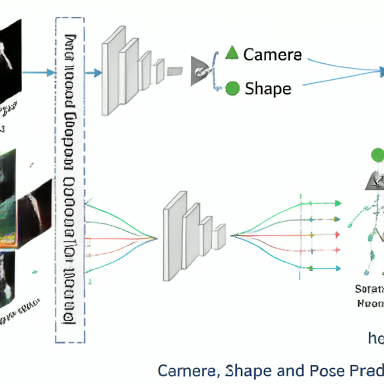}
                \\
                
                \includegraphics[width=0.19\linewidth]{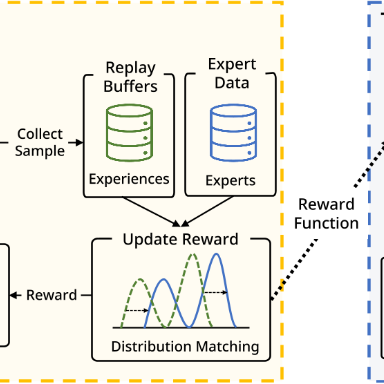} &
                \includegraphics[width=0.19\linewidth]{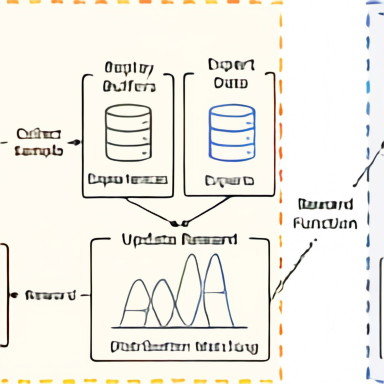} &
                \includegraphics[width=0.19\linewidth]{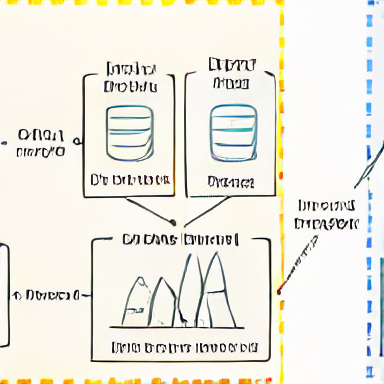} &
                \includegraphics[width=0.19\linewidth]{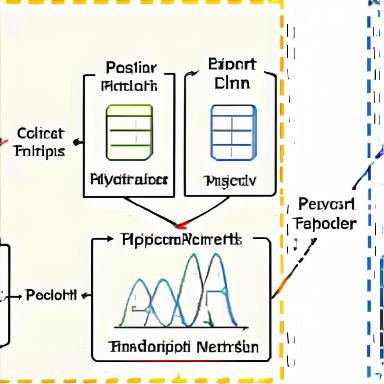} &
                \includegraphics[width=0.19\linewidth]{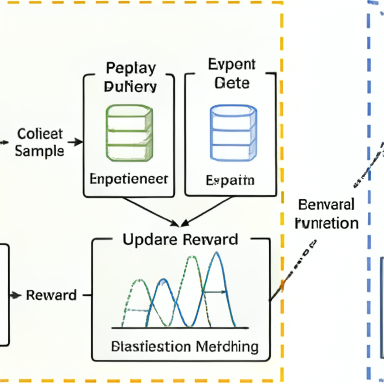}
                \\
                \midrule
                \multicolumn{5}{c}{\footnotesize\textbf{Evaluated on ICDAR13}}\\

                \includegraphics[width=0.19\linewidth]{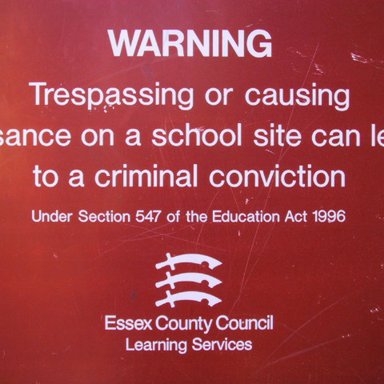} &
                \includegraphics[width=0.19\linewidth]{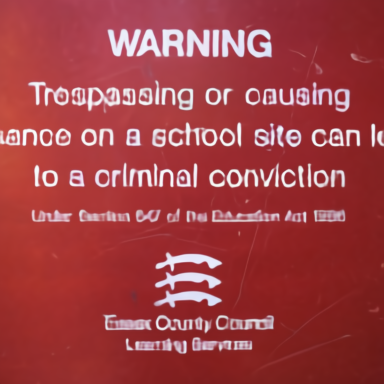} &
                \includegraphics[width=0.19\linewidth]{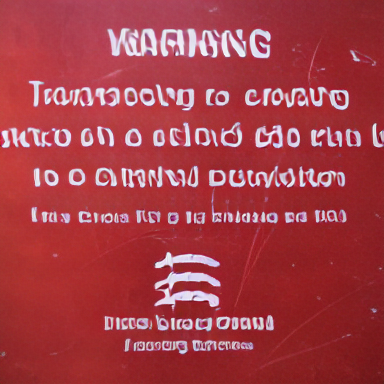} &
                \includegraphics[width=0.19\linewidth]{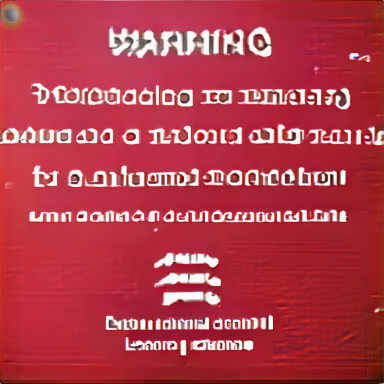} &
                \includegraphics[width=0.19\linewidth]{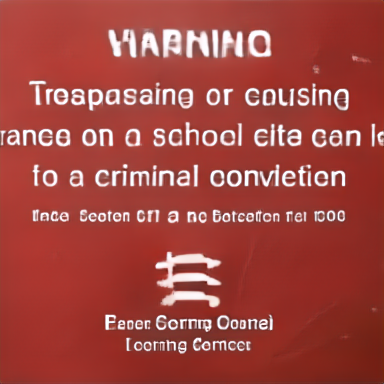}
                \\
                
                \includegraphics[width=0.19\linewidth]{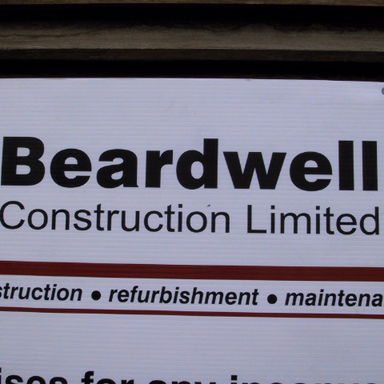} &
                \includegraphics[width=0.19\linewidth]{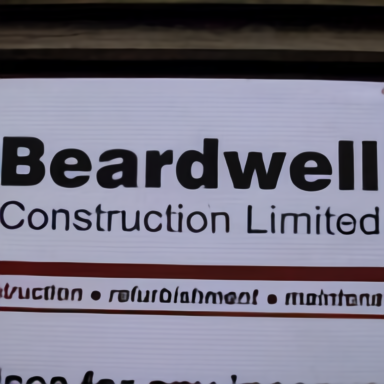} &
                \includegraphics[width=0.19\linewidth]{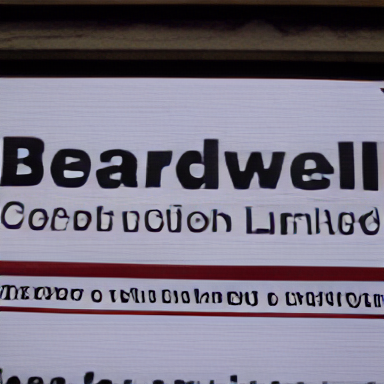} &
                \includegraphics[width=0.19\linewidth]{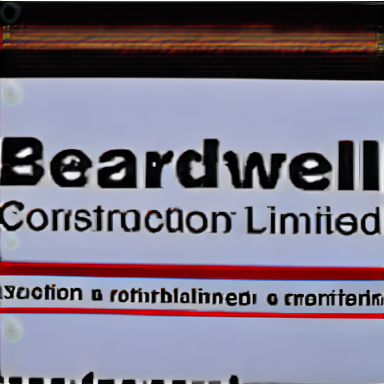} &
                \includegraphics[width=0.19\linewidth]{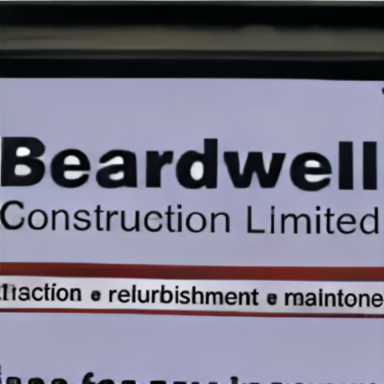}
                \\
                
                \includegraphics[width=0.19\linewidth]{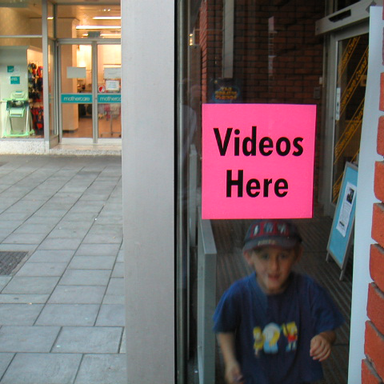} &
                \includegraphics[width=0.19\linewidth]{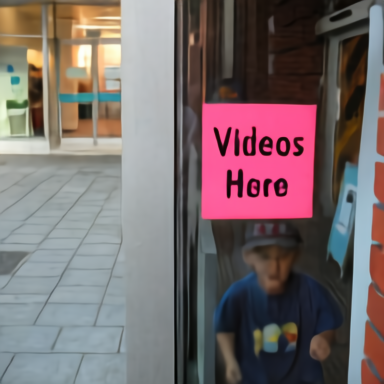} &
                \includegraphics[width=0.19\linewidth]{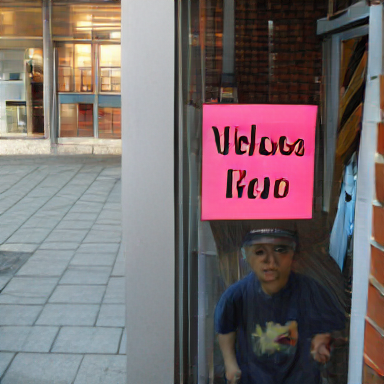} &
                \includegraphics[width=0.19\linewidth]{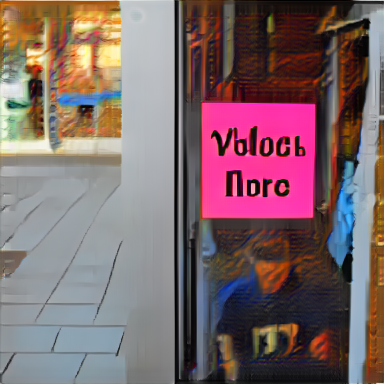} &
                \includegraphics[width=0.19\linewidth]{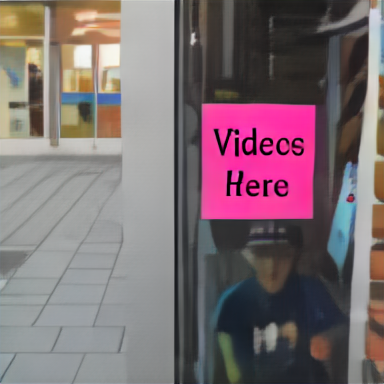}
                \\

                \bottomrule
            \end{tabular}
        }
    
    \end{center}
    \caption{Qualitative results on the reconstruction task of different image encoders. OCR-VQGAN outperforms other methods in figure-based images (Paper2Fig100k) both in the clarity of texts and diagrams. Results regarding ICDAR13 dataset (never seen in training) show that VQVAE gives similar results to OCR-VQGAN. It can be seen that OCR-VQGAN highlights text regions and adds ``figure style'' to natural images.}
    \label{table:method_comparison}
\end{table*}

\subsection{Evaluation Metrics}
We use image reconstruction-based metrics that allow us to measure the similarity between input and reconstructed images in terms of high-level semantics. To this end, distances in the feature space are more suitable.

FID~\citep{FID_paper,torchfidelity} is a reconstruction metric related to the diversity of the generated images with respect to the original ones. LPIPS~\citep{LPIPS} is a learned perceptual similarity metric. Both of these metrics measure appearance in reference to natural images, as they use features pre-trained on an Imagenet task. However, these metrics are insensitive to the text generation quality, which plays an important role in figures and diagrams. We introduce a third metric, OCR similarity (OCR-SIM), that quantifies the similarity of images that include text. This metric is computed as in equation \ref{eq:OCR_LOSS}, and it is more adequate to assess the proposed method, where the quality of the generated text is important.

\section{Experimental Results}
In this section, we conduct experiments to assess the performance of the proposed method in the task of figure reconstruction. Our goal is to obtain the optimal configuration of our method and compare OCR-VQGAN with popular image encoders using reconstruction metrics. 

\subsection{Training setting}
We use images of size $384\times384$, which we empirically set to maximize resolution and GPU memory. Images are resized to the smallest size between $H$ and $W$, and randomly cropped. The baseline VQGAN architecture has 112M parameters, a codebook of size $16,384$ with embeddings vectors of size $256$, and it is pre-trained on Imagenet (VQGAN$_{Imagenet}$). VQGAN encodes images with a downsampling factor of 16, resulting in grids of $24\times24$ (or sequences of 576 tokens). We also use a pre-trained VQVAE model from DALLE~\citep{DALLE}, which consists of a codebook of size 8192.

For training the models, we use data parallelism with 4 V100 GPUs with a total effective batch size of $16$ for $20$ epochs. We perform an initial warm-up that does not use the discriminator, as has been empirically found beneficial for better reconstructions~\citep{VQGAN}. We use the Adam optimizer with a learning rate of $4.5
\times10^{-4}$. 

\subsection{Training datasets}
We train OCR-VQGAN using two datasets of images that contain text. \textbf{Paper2Fig100k}, presented in Section~\ref{sec:Paper2Fig100k}, contains $81,194$ training samples and $21,259$ test samples of figure diagrams with rendered text. \textbf{ICDAR13} \citep{ICDAR13}, was presented during the ICDAR 2013 Robust Reading Competition, focused on the task of scene text detection. The dataset is composed of high-resolution natural images with texts in English to test methods when texts are displayed in natural scenes. The dataset contains $229$ samples for train and $233$ for testing, but we use all $462$ samples as a test set, as we use this dataset only for evaluation. 

\subsection{Complementary Perceptual Losses}
We perform a hyper-parameter search for OCR-VQGAN over different weighting configurations of the LPIPS and OCR perceptual losses. Specifically, we scale the two losses using weights $w_{ocr}$ and $w_{vgg}$. Results in Table~\ref{table:results_p_losses} show that $w_{ocr}$ is more important than $w_{vgg}$ when dealing with figures and diagrams. The use of a small weight for $w_{vgg}$ (between 0.2 and 0.5) gives the best performance. This is because some figures present small regions with natural images. In those regions, the LPIPS features get more activated, and the loss improves.

\paragraph{OCR model overhead.}
Table~\ref{table:overhead} reports results on the performance overhead of adding the OCR model to VQGAN, in terms of network parameters and test time. We also show the parameters that define the latent space $\mathcal{Z}$. Test time is measured during the evaluation of Paper2Fig100k test dataset. We also evaluate test speed on the VQVAE model from DALLE~\cite{DALLE}, which uses a smaller discrete latent space. This result shows an acceptable increment of test time given the increase in parameters (20M) and the gain in qualitative performance.

\subsection{Evaluation of image tokenizers}
We analyze model performance on the tasks of figure-based reconstruction (Paper2Fig100k) and natural text-based reconstruction (ICDAR13) (Tables~\ref{table:results_method_comparison} and \ref{table:method_comparison}). The proposed OCR-VQGAN model outperforms the other methods both quantitatively and qualitatively, being able to display almost all details in the figures. One limitation is that text can only be recovered when it is sufficiently large. This can be solved using larger resolutions or upscaling models. We found that vertical text is also challenging to reconstruct as well as uncommon background colors. 

Results on IDCAR13 dataset show acceptable LPIPS and OCR-SIM, even though the models were not trained on that dataset. VQVAE and VQGAN show better FID scores because they were trained using natural images. VQVAE displays most of the natural text but fails with small text sizes. OCR-VQGAN, fine-tuned with figures, reconstructs appealing natural images and displays texts. It also reconstructs images with its own ``figure style'' by smoothing textures and highlighting the rendered texts. As expected, its main limitation is that it fails when text appears small, with orientation, and with infrequent color and background combinations (see the ``Warning'' example from Table~\ref{table:results_method_comparison}).

\section{Conclusion}
We focused on generating diagrams with clear texts within images. We proposed OCR-VQGAN as an image encoder and decoder to improve within-image text generation. We add a loss term for OCR perceptual similarity as a complement to the default VGG LPIPS to a VQGAN architecture. In addition, we presented Paper2Fig100k, the first text-to-image dataset in the domain of research papers and figures. We conducted several experiments that demonstrate how the OCR perceptual loss is beneficial for generating clear texts and diagram shapes. Results show that a small weight for the VGG term is also beneficial. We hope our work constitutes a first stepping stone towards text-to-figure generation.

\section{Ethics and social impact}
In this work, we focus on generating paper figures and argue how this application could be useful for researchers in the process of generating understandable diagrams, and potentially help a broad audience when creating appealing and effective slide presentations. However, a central concern of this system is that it could be used for fake paper generation and for bypassing plagiarism detection systems. 

Some steps that can be made to address this ethical concern is to build classifiers that allow for the detection of fake or plagiarized content. Experiments can be done in order to use the acquired knowledge learned by text-to-image models to train a discriminator. In the Parti paper~\citep{Parti}, authors propose the use of watermarks in the generated images, in order to easily detect AI-generated samples. Also, the  proposed dataset can be leveraged to train plagiarism detection systems for research publications.

Further research is needed to elucidate how these systems should be made public in order to align their behavior with ethical standards.

\clearpage
{\small
\bibliographystyle{unsrtnat}
\bibliography{mainbib}
}
\clearpage
\begin{appendices}
\section{Additional qualitative results}
In this section, we give additional qualitative results on both Paper2Fig100k and ICDAR13 datasets. We perform a random sampling of the generated images in the test sets and display the comparison with different methods. We aim at finding the limitations of the proposed image encoder when rendering texts within figures and text-in-the-wild.

Regarding results on Paper2Fig100k dataset, shown in tables~ \ref{table:appendix_Paper2Fig100k_table1} and \ref{table:appendix_Paper2Fig100k_table2}, OCR-VQGAN outperforms the other methods in almost all scenarios (challenging shapes, text sizes, orientations, and colors). It gives qualitatively better results concerning both VQGAN alternatives (Imagenet pre-trained and Paper2Fig100k finetuned), showing that the OCR loss is beneficial when reconstructing text-within-figures. VQVAE, trained for DALLE, gives acceptable results when conditions are favorable, such as having a simple background-color combination or when texts have sufficiently large sizes. With long words or sentences, OCR-VQGAN can display sharper characters, whereas other methods tend to merge them. Arrows, straight lines, or dashed lines are sharper in OCR-VQGAN. We can observe a trade-off between the blurriness and clearness of text in VQVAE and VQGAN, where VQVAE generates more blurry samples. The limitations of our method are shown when complex color-background combinations are present, when the text is very small (low resolution) and when the text is displayed in a vertical orientation.

Qualitative results on ICDAR13 dataset are presented in tables~ \ref{table:appendix_ICDAR2013_table1} and \ref{table:appendix_ICDAR2013_table2}. None of the methods were trained using these images, therefore the goal is to test how they perform in never-seen images of text-in-the-wild. The proposed OCR-VQGAN, even though it was trained with 80k images of figures, can reconstruct appealing natural images with in-the-wild texts. As shown in most of the samples, it is transferring the style of figures, tending to smooth out the textures, and highlighting the texts. However, it is sensitive to complex lighting, textures, and image quality conditions. Text is not readable in some cases. VQGAN approaches fail at the task of text reconstruction, where texts are mostly unreadable. VQVAE gives good reconstruction results in ICDAR13, generating natural-looking images and mostly readable text. This is because VQVAE was trained using natural images and it can handle challenging in-the-wild conditions. OCR-VQGAN tends to focus the attention on the texts, while VQVAE gives a more smooth generation in ICDAR13. 

\begin{table*}[t]
    \begin{center}
    \setlength{\tabcolsep}{2pt}
        \resizebox{0.95\textwidth}{!}{
            \begin{tabular}{ccccc}
                \toprule
               \textbf{Ground truth} & \textbf{VQVAE$_{\text{\textit{DALLE}}}$} & \textbf{VQGAN$_{\text{\textit{Imagenet}}}$} & \textbf{VQGAN$_{\text{\textit{Paper2Fig100k}}}$} & \textbf{OCR-VQGAN} \\
                \midrule
                \multicolumn{5}{c}{\footnotesize\textbf{Evaluated on Paper2Fig100k}}\\
                
                \includegraphics[width=0.19\linewidth]{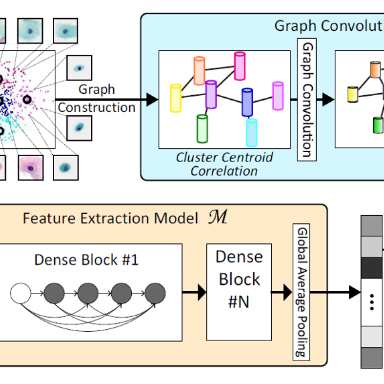} &
                \includegraphics[width=0.19\linewidth]{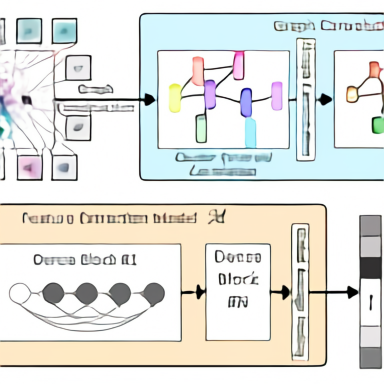} &
                \includegraphics[width=0.19\linewidth]{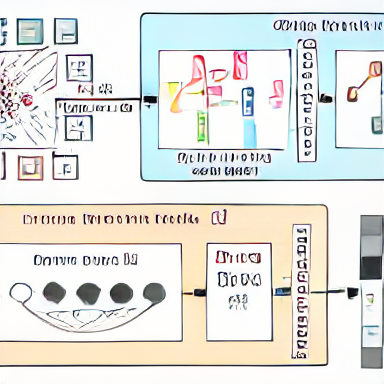} &
                \includegraphics[width=0.19\linewidth]{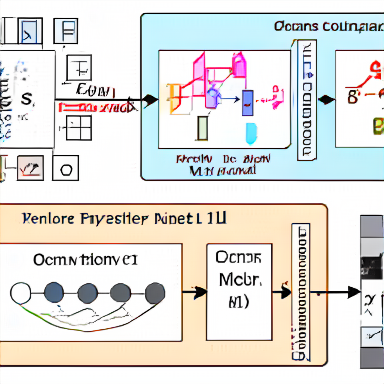} &
                \includegraphics[width=0.19\linewidth]{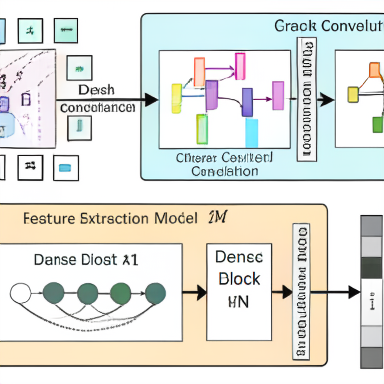}
                \\
                
                \includegraphics[width=0.19\linewidth]{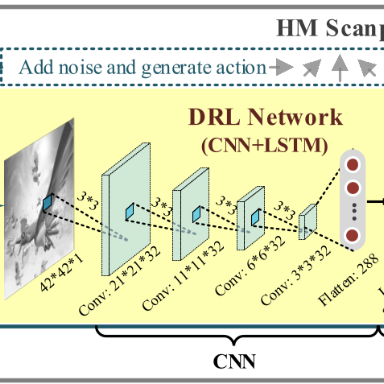} &
                \includegraphics[width=0.19\linewidth]{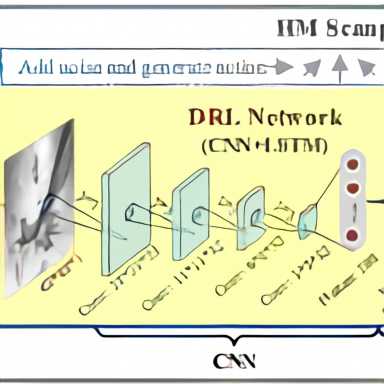} &
                \includegraphics[width=0.19\linewidth]{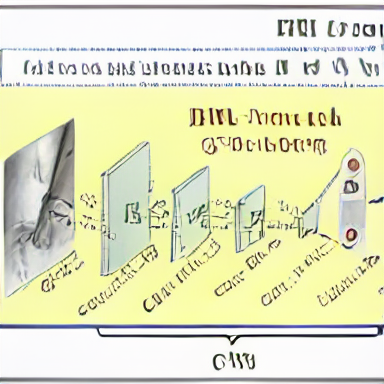} &
                \includegraphics[width=0.19\linewidth]{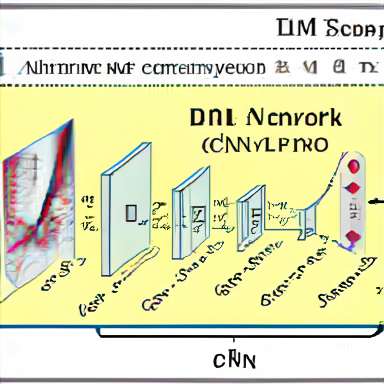} &
                \includegraphics[width=0.19\linewidth]{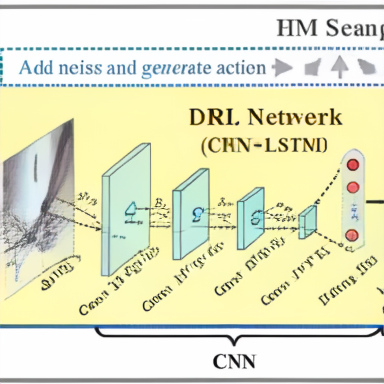}
                \\
                
                \includegraphics[width=0.19\linewidth]{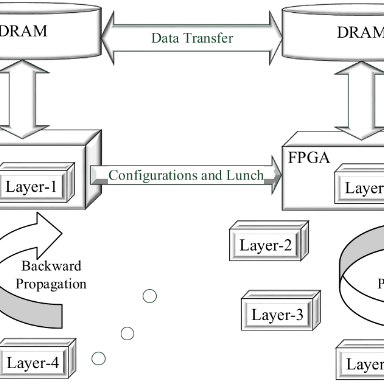} &
                \includegraphics[width=0.19\linewidth]{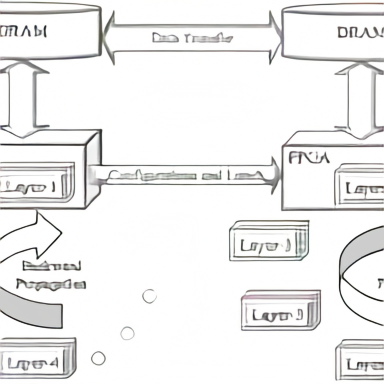} &
                \includegraphics[width=0.19\linewidth]{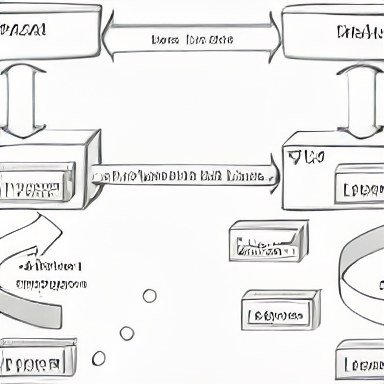} &
                \includegraphics[width=0.19\linewidth]{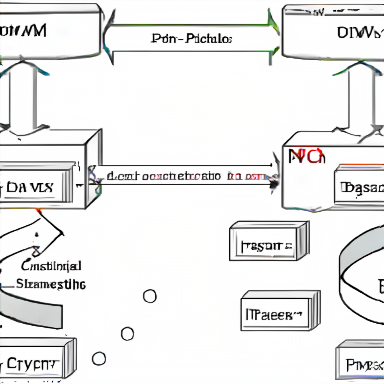} &
                \includegraphics[width=0.19\linewidth]{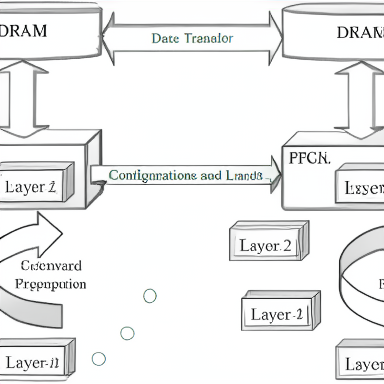}
                \\
                 
                \includegraphics[width=0.19\linewidth]{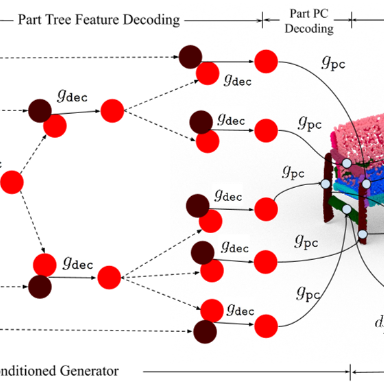} &
                \includegraphics[width=0.19\linewidth]{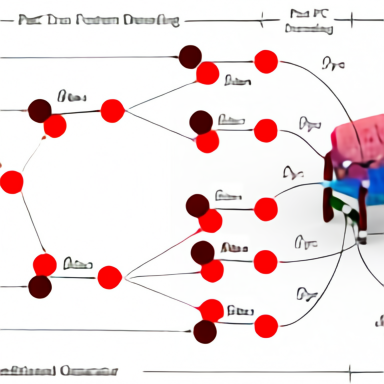} &
                \includegraphics[width=0.19\linewidth]{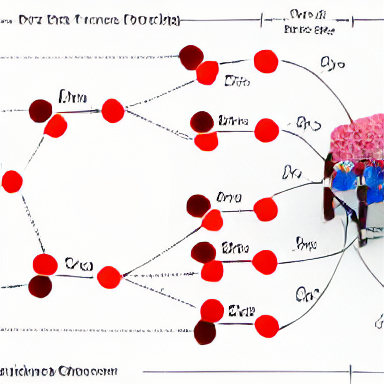} &
                \includegraphics[width=0.19\linewidth]{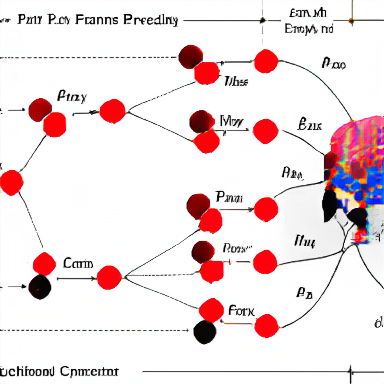} &
                \includegraphics[width=0.19\linewidth]{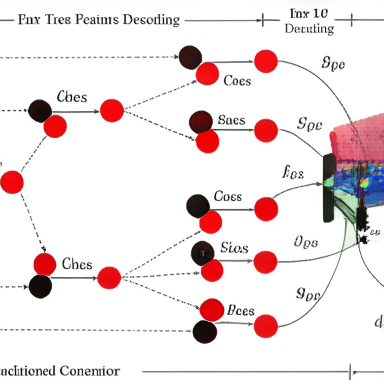}
                \\
                
                \includegraphics[width=0.19\linewidth]{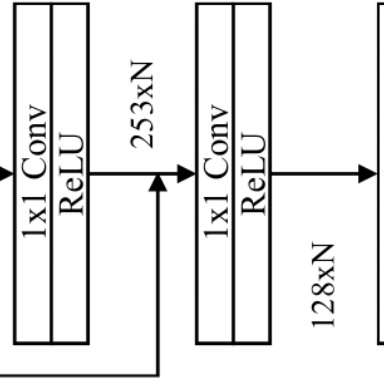} &
                \includegraphics[width=0.19\linewidth]{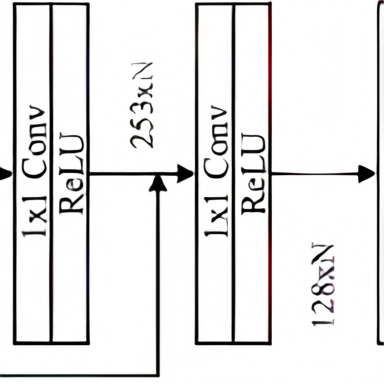} &
                \includegraphics[width=0.19\linewidth]{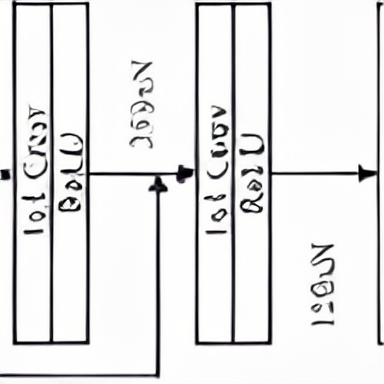} &
                \includegraphics[width=0.19\linewidth]{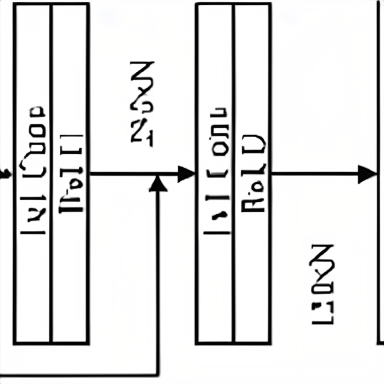} &
                \includegraphics[width=0.19\linewidth]{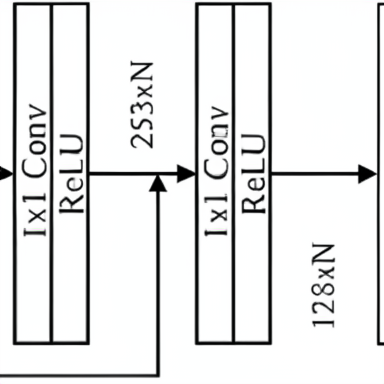}
                \\
                
                \includegraphics[width=0.19\linewidth]{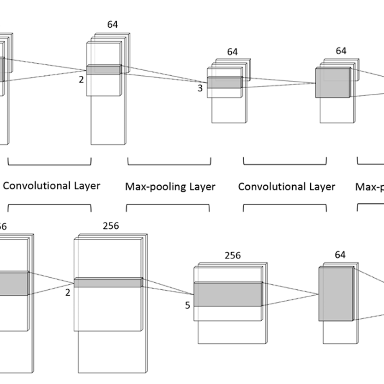} &
                \includegraphics[width=0.19\linewidth]{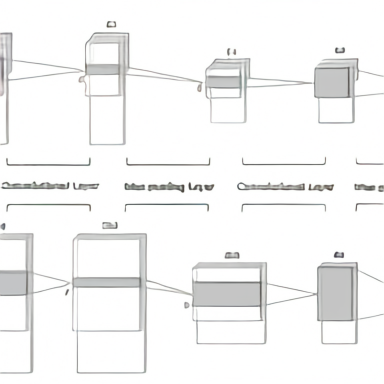} &
                \includegraphics[width=0.19\linewidth]{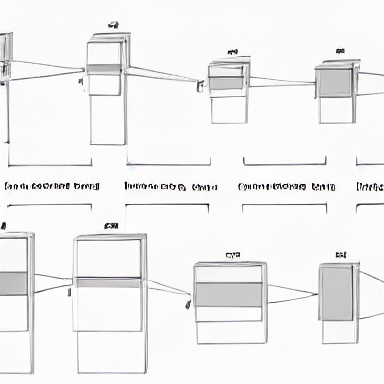} &
                \includegraphics[width=0.19\linewidth]{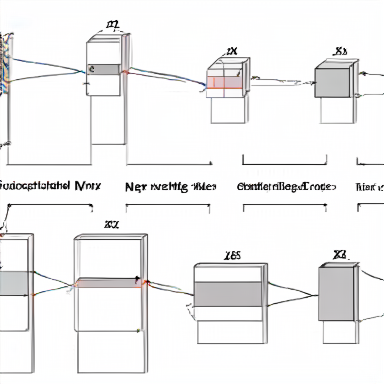} &
                \includegraphics[width=0.19\linewidth]{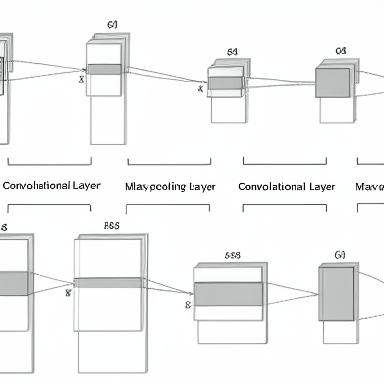}
                \\

                \bottomrule
            \end{tabular}
        }
    
    \end{center}
    \caption{Reconstructed images from Paper2Fig100k test set.}
    \label{table:appendix_Paper2Fig100k_table1}
\end{table*}

\begin{table*}[t]
    \begin{center}
    \setlength{\tabcolsep}{2pt}
        \resizebox{0.95\textwidth}{!}{
            \begin{tabular}{ccccc}
                \toprule
               \textbf{Ground truth} & \textbf{VQVAE$_{\text{\textit{DALLE}}}$} & \textbf{VQGAN$_{\text{\textit{Imagenet}}}$} & \textbf{VQGAN$_{\text{\textit{Paper2Fig100k}}}$} & \textbf{OCR-VQGAN} \\
                \midrule
                \multicolumn{5}{c}{\footnotesize\textbf{Evaluated on Paper2Fig100k}}\\

                \includegraphics[width=0.19\linewidth]{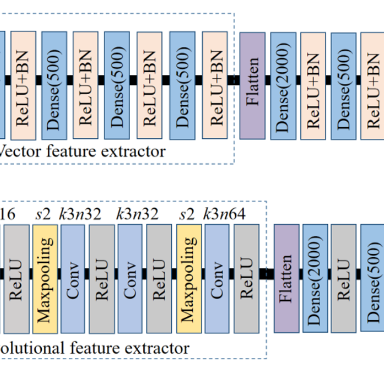} &
                \includegraphics[width=0.19\linewidth]{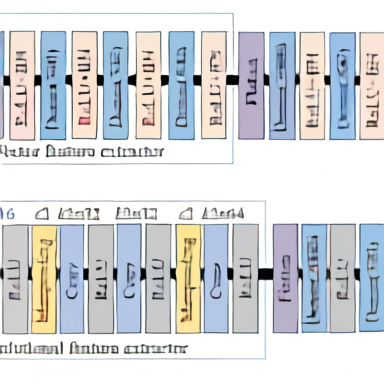} &
                \includegraphics[width=0.19\linewidth]{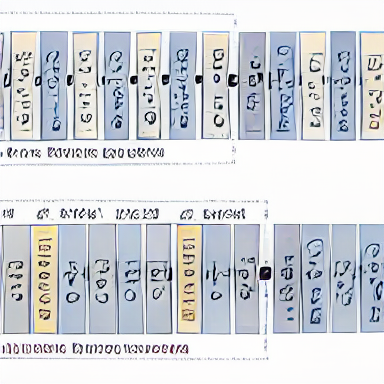} &
                \includegraphics[width=0.19\linewidth]{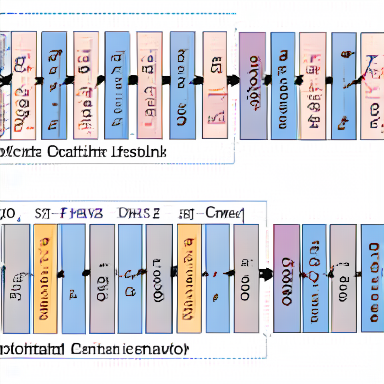} &
                \includegraphics[width=0.19\linewidth]{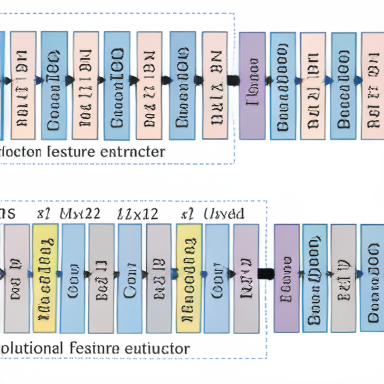}
                \\
                 
                \includegraphics[width=0.19\linewidth]{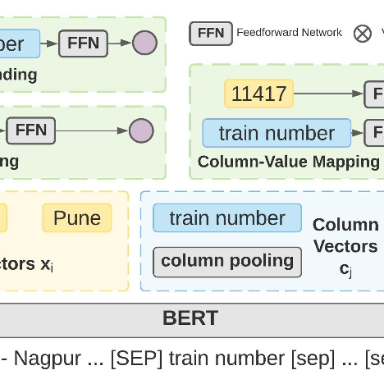} &
                \includegraphics[width=0.19\linewidth]{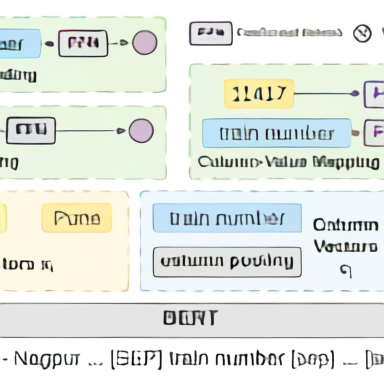} &
                \includegraphics[width=0.19\linewidth]{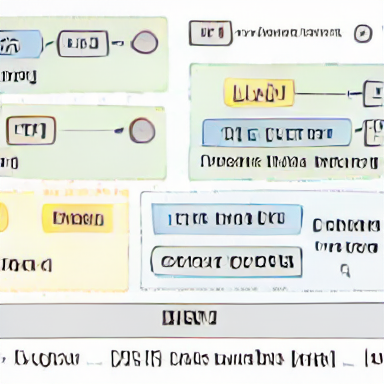} &
                \includegraphics[width=0.19\linewidth]{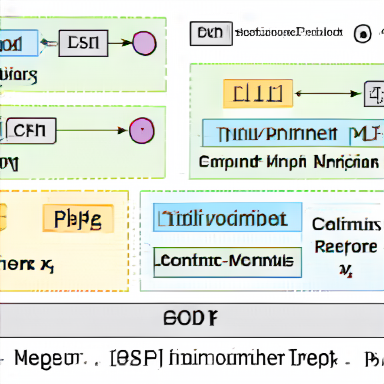} &
                \includegraphics[width=0.19\linewidth]{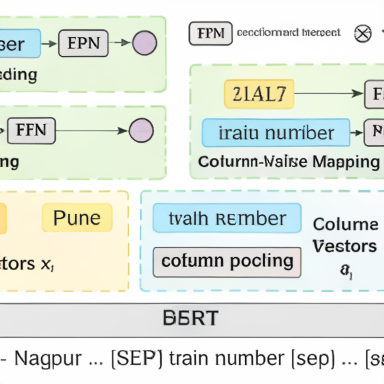}
                \\
                
                \includegraphics[width=0.19\linewidth]{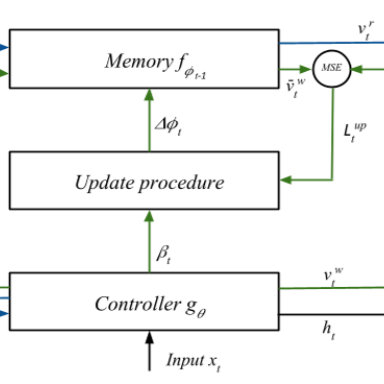} &
                \includegraphics[width=0.19\linewidth]{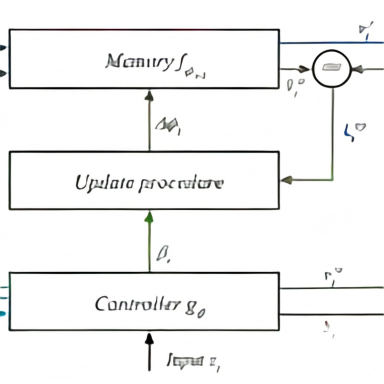} &
                \includegraphics[width=0.19\linewidth]{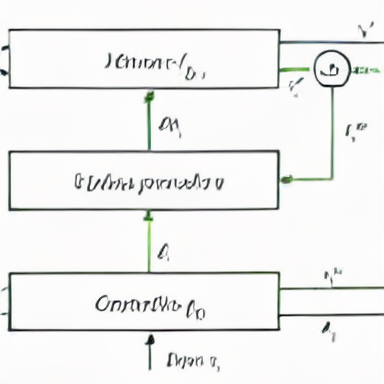} &
                \includegraphics[width=0.19\linewidth]{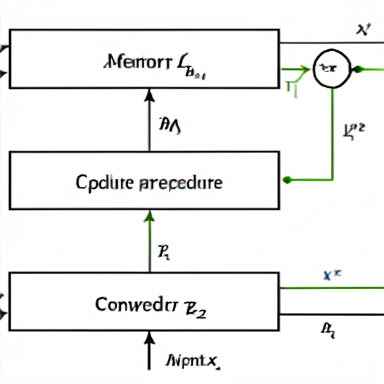} &
                \includegraphics[width=0.19\linewidth]{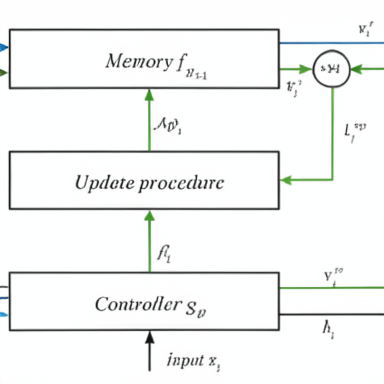}
                \\
                
                \includegraphics[width=0.19\linewidth]{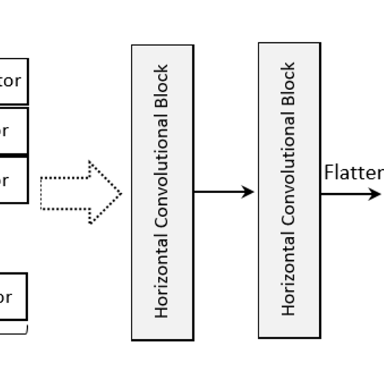} &
                \includegraphics[width=0.19\linewidth]{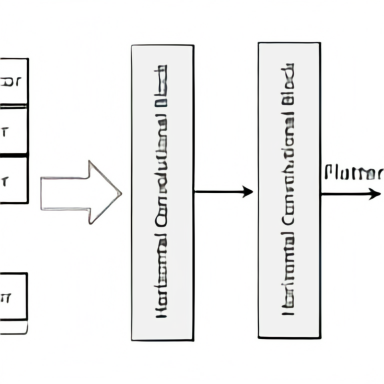} &
                \includegraphics[width=0.19\linewidth]{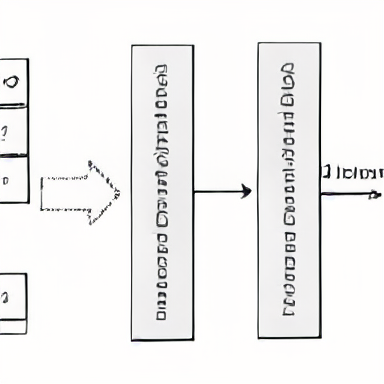} &
                \includegraphics[width=0.19\linewidth]{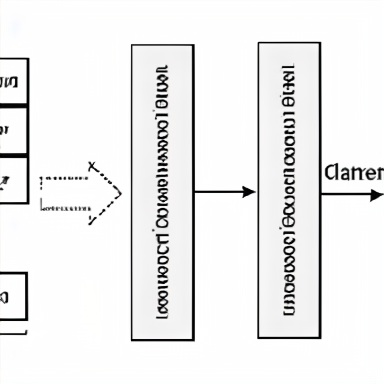} &
                \includegraphics[width=0.19\linewidth]{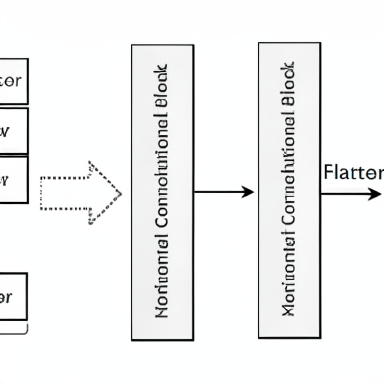}
                \\
                
                \includegraphics[width=0.19\linewidth]{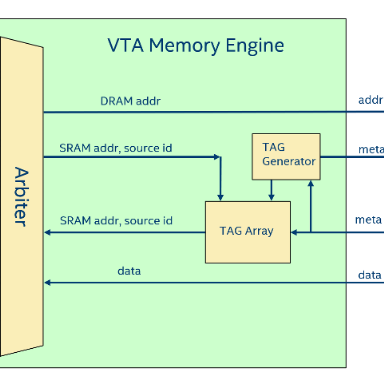} &
                \includegraphics[width=0.19\linewidth]{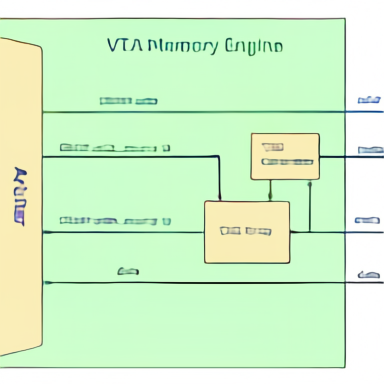} &
                \includegraphics[width=0.19\linewidth]{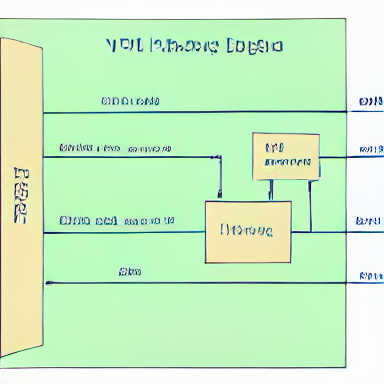} &
                \includegraphics[width=0.19\linewidth]{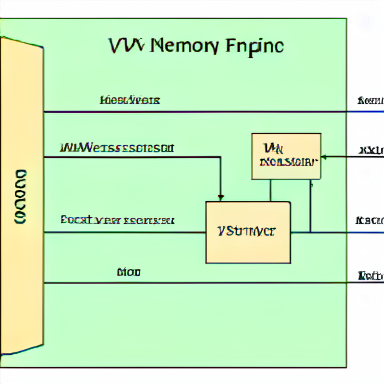} &
                \includegraphics[width=0.19\linewidth]{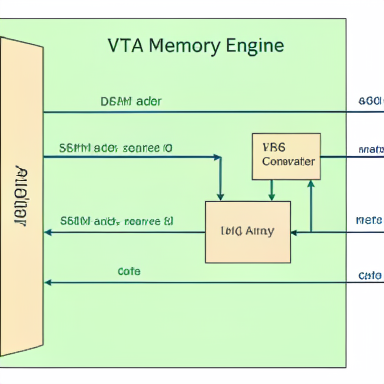}
                \\
                
                \includegraphics[width=0.19\linewidth]{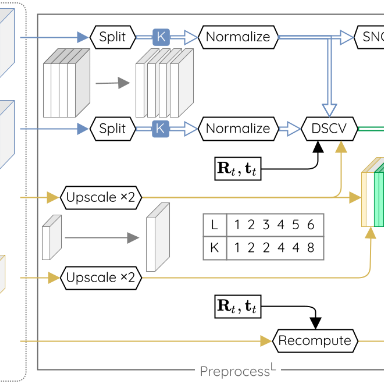} &
                \includegraphics[width=0.19\linewidth]{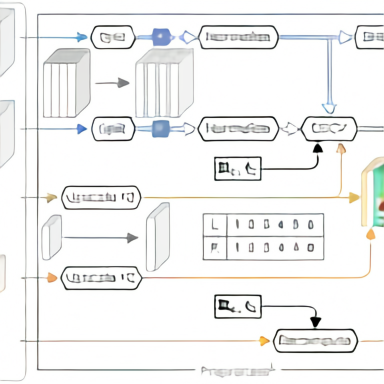} &
                \includegraphics[width=0.19\linewidth]{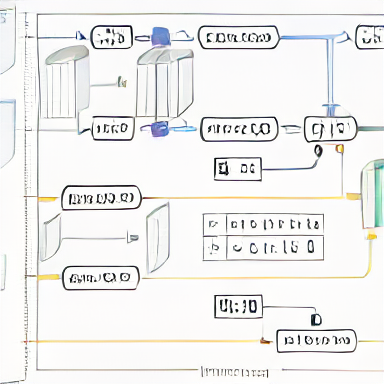} &
                \includegraphics[width=0.19\linewidth]{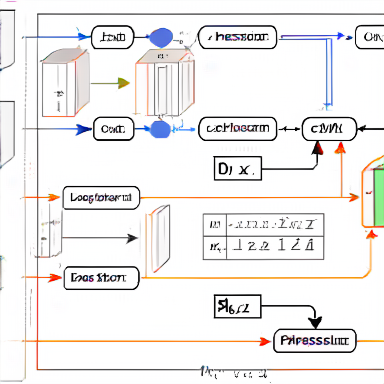} &
                \includegraphics[width=0.19\linewidth]{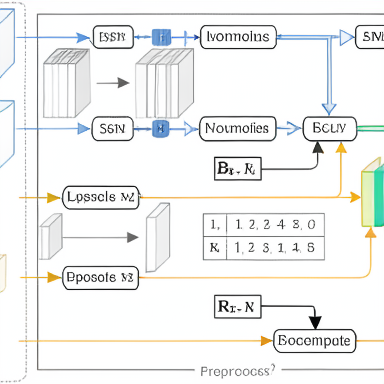}
                \\
                
                \bottomrule
            \end{tabular}
        }
    
    \end{center}
    \caption{Reconstructed images from Paper2Fig100k test set.}
    \label{table:appendix_Paper2Fig100k_table2}
\end{table*}

\begin{table*}[t]
    \begin{center}
    \setlength{\tabcolsep}{2pt}
        \resizebox{0.95\textwidth}{!}{
            \begin{tabular}{ccccc}
                \toprule
               \textbf{Ground truth} & \textbf{VQVAE$_{\text{\textit{DALLE}}}$} & \textbf{VQGAN$_{\text{\textit{Imagenet}}}$} & \textbf{VQGAN$_{\text{\textit{Paper2Fig100k}}}$} & \textbf{OCR-VQGAN} \\
                \midrule
                \multicolumn{5}{c}{\footnotesize\textbf{Evaluated on ICDAR13}}\\
                
                \includegraphics[width=0.19\linewidth]{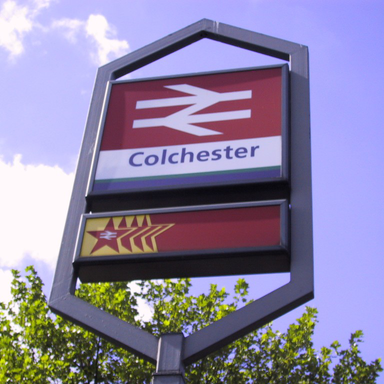} &
                \includegraphics[width=0.19\linewidth]{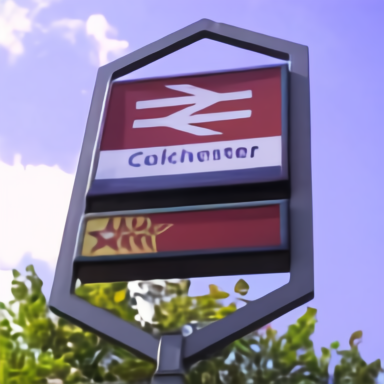} &
                \includegraphics[width=0.19\linewidth]{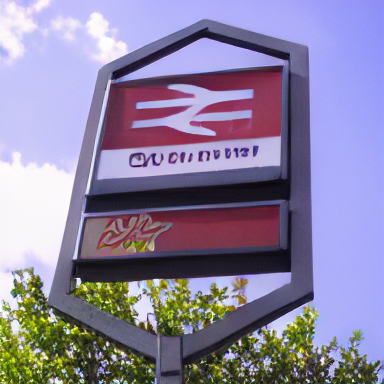} &
                \includegraphics[width=0.19\linewidth]{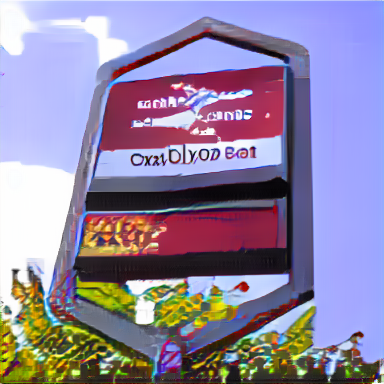} &
                \includegraphics[width=0.19\linewidth]{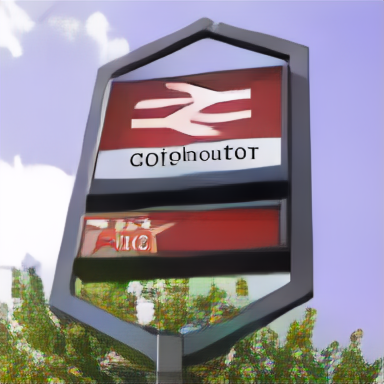}
                \\
                
                \includegraphics[width=0.19\linewidth]{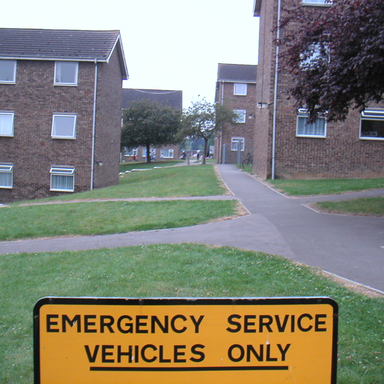} &
                \includegraphics[width=0.19\linewidth]{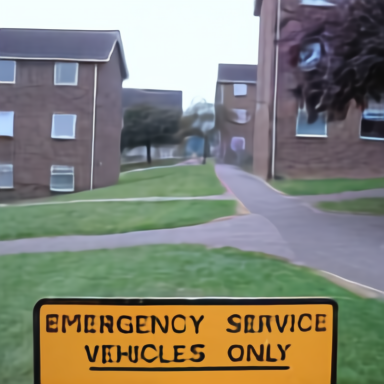} &
                \includegraphics[width=0.19\linewidth]{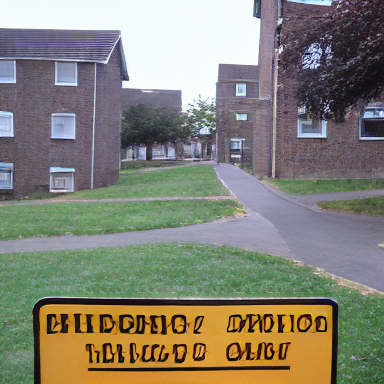} &
                \includegraphics[width=0.19\linewidth]{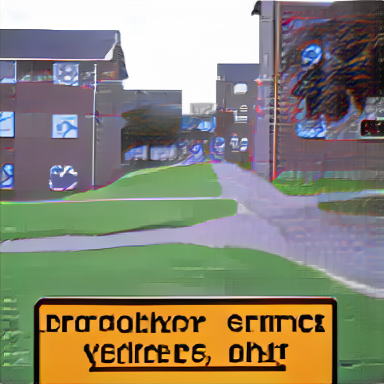} &
                \includegraphics[width=0.19\linewidth]{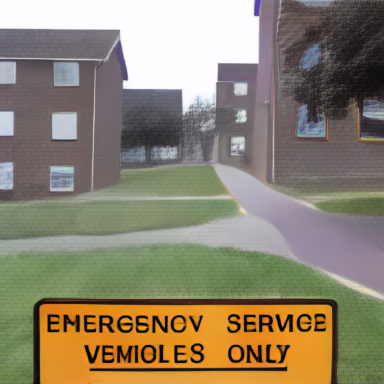}
                \\
                
                \includegraphics[width=0.19\linewidth]{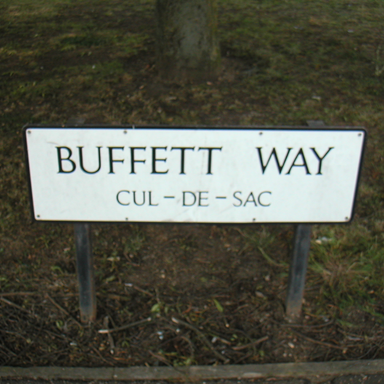} &
                \includegraphics[width=0.19\linewidth]{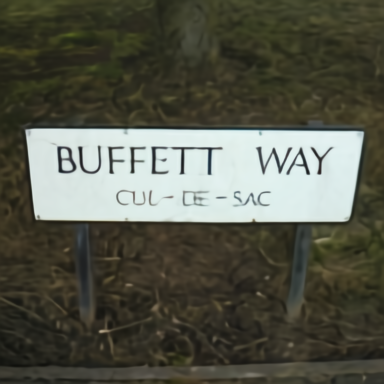} &
                \includegraphics[width=0.19\linewidth]{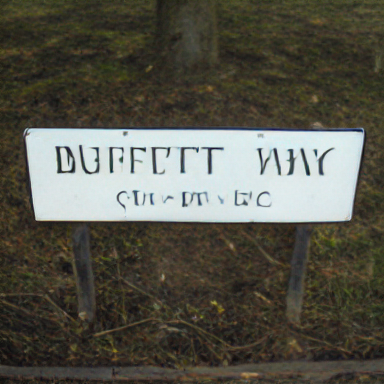} &
                \includegraphics[width=0.19\linewidth]{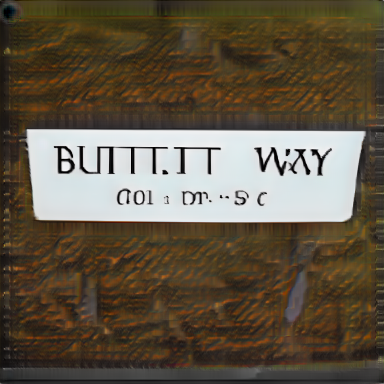} &
                \includegraphics[width=0.19\linewidth]{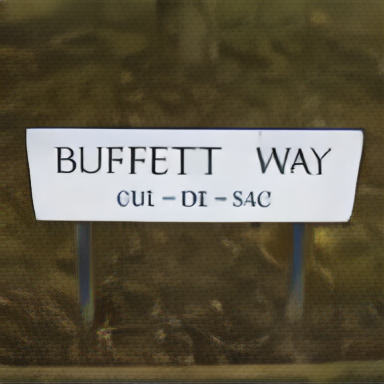}
                \\
                 
                \includegraphics[width=0.19\linewidth]{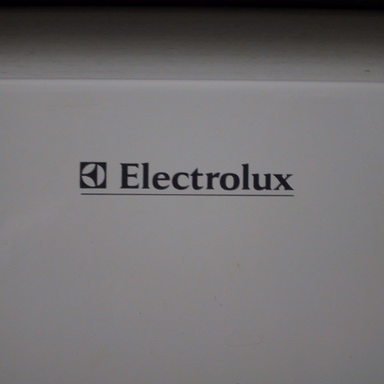} &
                \includegraphics[width=0.19\linewidth]{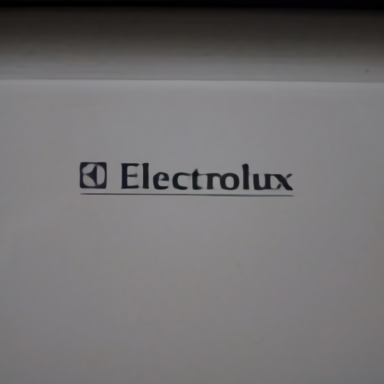} &
                \includegraphics[width=0.19\linewidth]{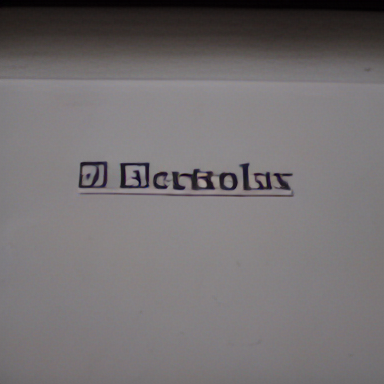} &
                \includegraphics[width=0.19\linewidth]{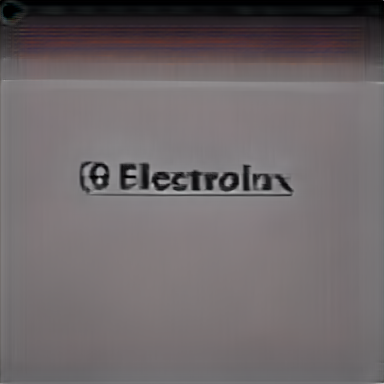} &
                \includegraphics[width=0.19\linewidth]{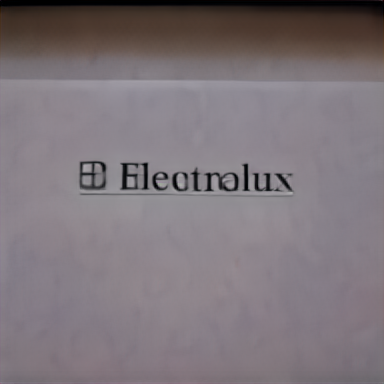}
                \\
                
                \includegraphics[width=0.19\linewidth]{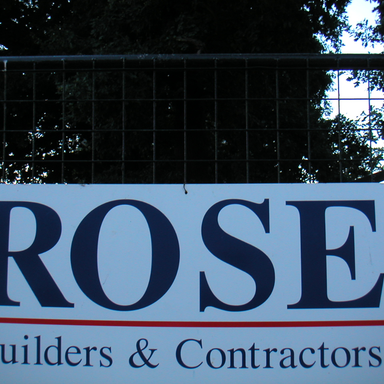} &
                \includegraphics[width=0.19\linewidth]{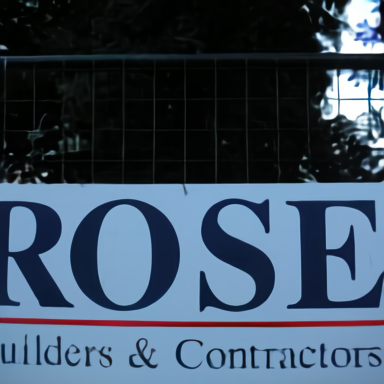} &
                \includegraphics[width=0.19\linewidth]{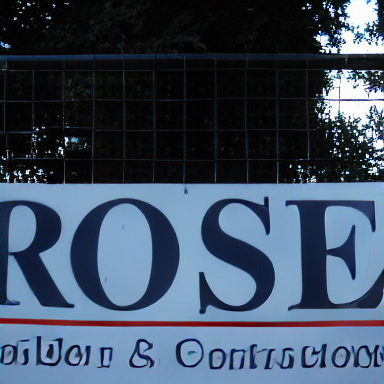} &
                \includegraphics[width=0.19\linewidth]{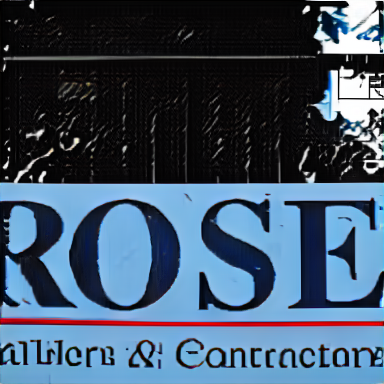} &
                \includegraphics[width=0.19\linewidth]{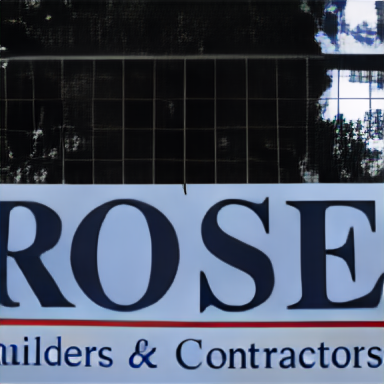}
                \\
                
                \includegraphics[width=0.19\linewidth]{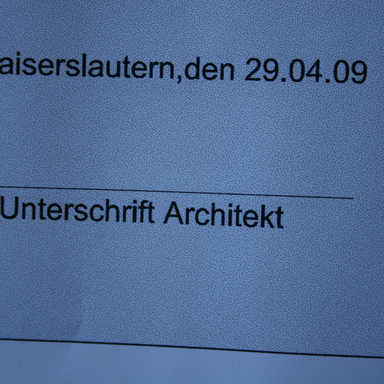} &
                \includegraphics[width=0.19\linewidth]{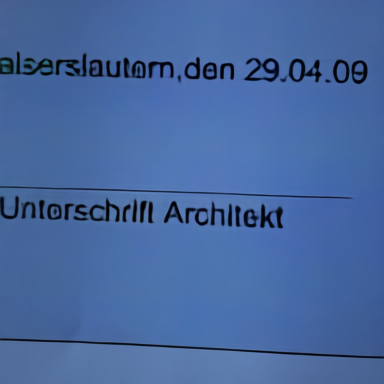} &
                \includegraphics[width=0.19\linewidth]{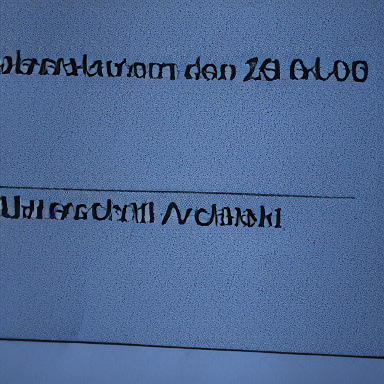} &
                \includegraphics[width=0.19\linewidth]{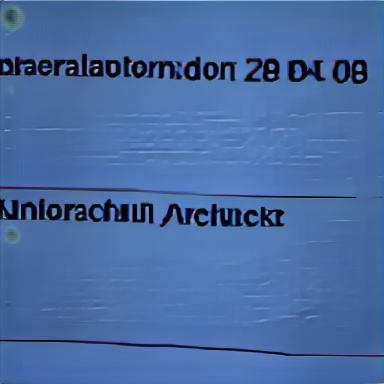} &
                \includegraphics[width=0.19\linewidth]{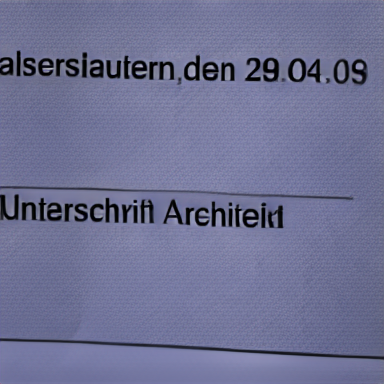}
                \\

                \bottomrule
            \end{tabular}
        }
    
    \end{center}
    \caption{Reconstructed images from ICDAR13 test set.}
    \label{table:appendix_ICDAR2013_table1}
\end{table*}

\begin{table*}[t]
    \begin{center}
    \setlength{\tabcolsep}{2pt}
        \resizebox{0.95\textwidth}{!}{
            \begin{tabular}{ccccc}
                \toprule
               \textbf{Ground truth} & \textbf{VQVAE$_{\text{\textit{DALLE}}}$} & \textbf{VQGAN$_{\text{\textit{Imagenet}}}$} & \textbf{VQGAN$_{\text{\textit{Paper2Fig100k}}}$} & \textbf{OCR-VQGAN} \\
                \midrule
                \multicolumn{5}{c}{\footnotesize\textbf{Evaluated on ICDAR13}}\\

                \includegraphics[width=0.19\linewidth]{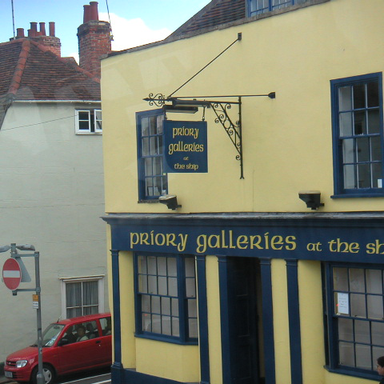} &
                \includegraphics[width=0.19\linewidth]{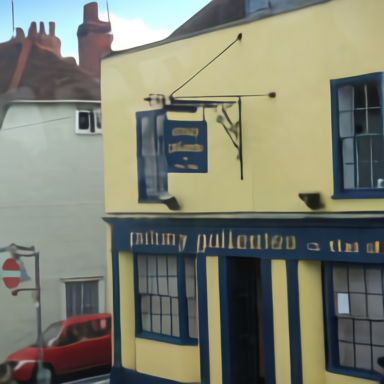} &
                \includegraphics[width=0.19\linewidth]{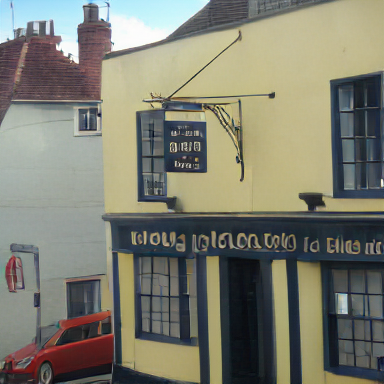} &
                \includegraphics[width=0.19\linewidth]{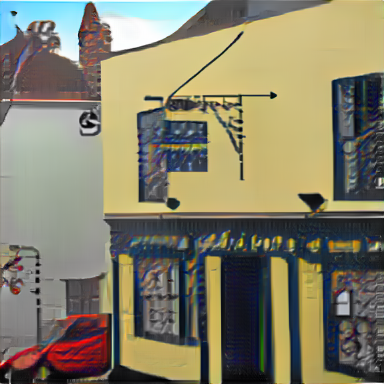} &
                \includegraphics[width=0.19\linewidth]{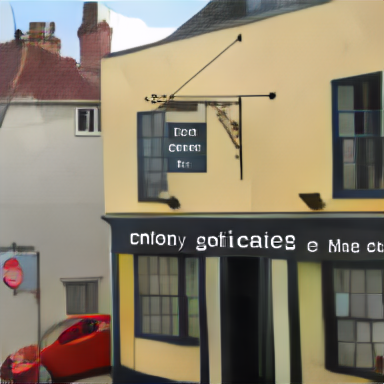}
                \\
                 
                \includegraphics[width=0.19\linewidth]{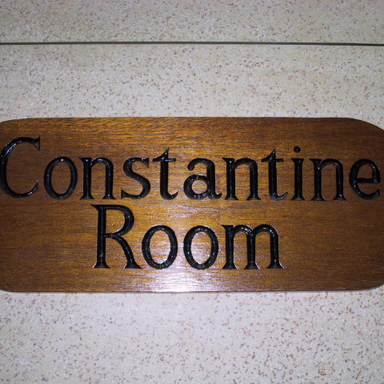} &
                \includegraphics[width=0.19\linewidth]{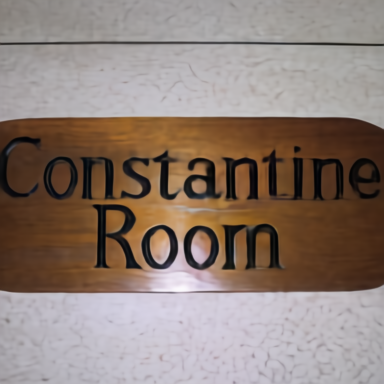} &
                \includegraphics[width=0.19\linewidth]{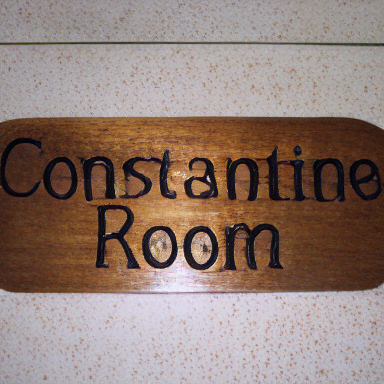} &
                \includegraphics[width=0.19\linewidth]{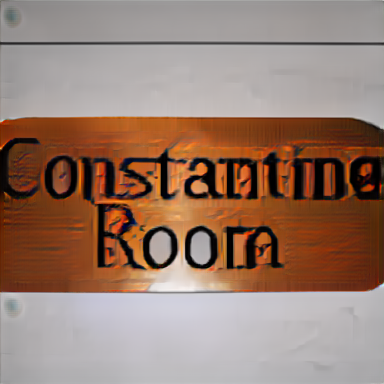} &
                \includegraphics[width=0.19\linewidth]{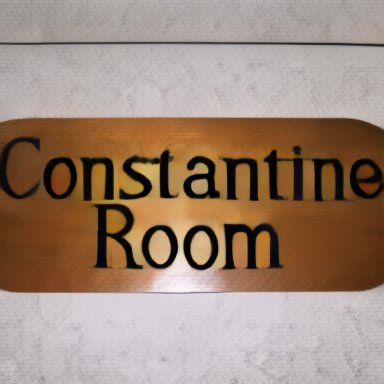}
                \\
                
                \includegraphics[width=0.19\linewidth]{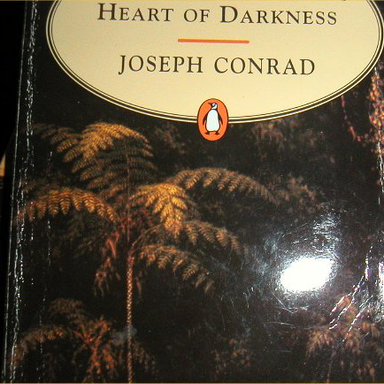} &
                \includegraphics[width=0.19\linewidth]{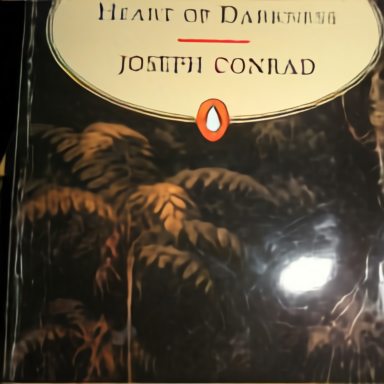} &
                \includegraphics[width=0.19\linewidth]{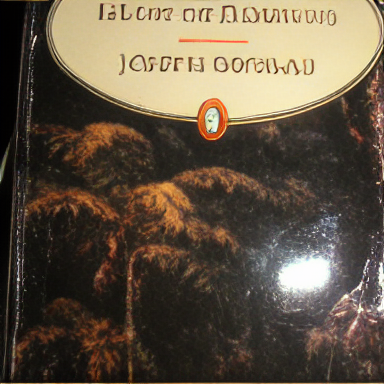} &
                \includegraphics[width=0.19\linewidth]{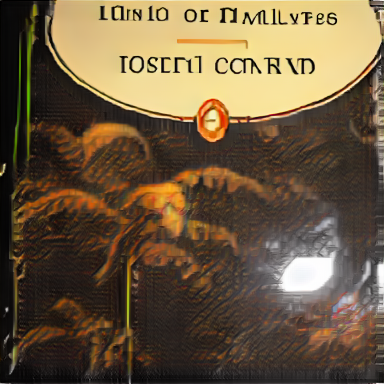} &
                \includegraphics[width=0.19\linewidth]{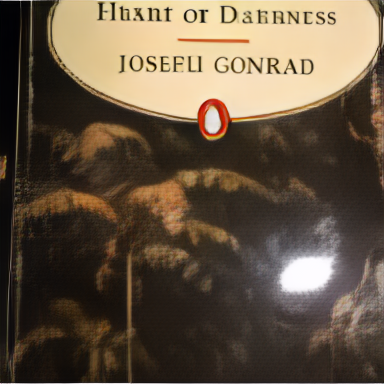}
                \\
                
                \includegraphics[width=0.19\linewidth]{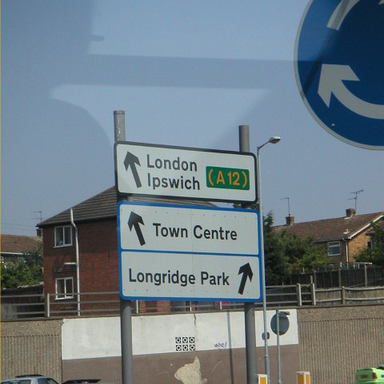} &
                \includegraphics[width=0.19\linewidth]{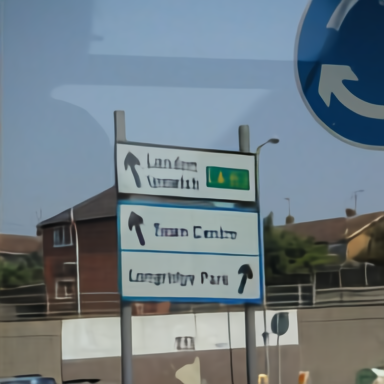} &
                \includegraphics[width=0.19\linewidth]{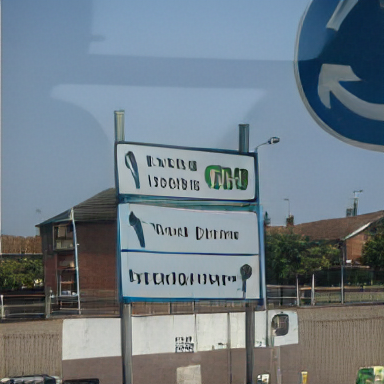} &
                \includegraphics[width=0.19\linewidth]{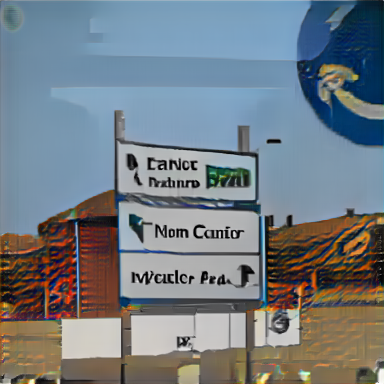} &
                \includegraphics[width=0.19\linewidth]{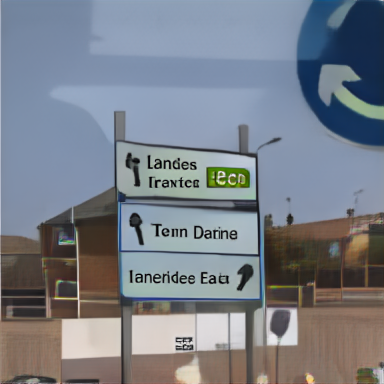}
                \\
                
                \includegraphics[width=0.19\linewidth]{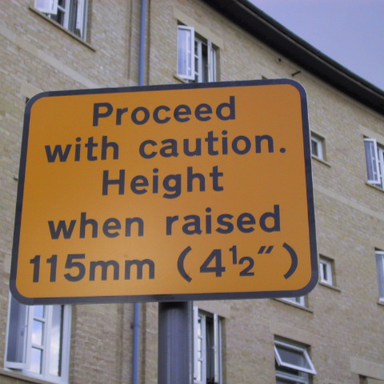} &
                \includegraphics[width=0.19\linewidth]{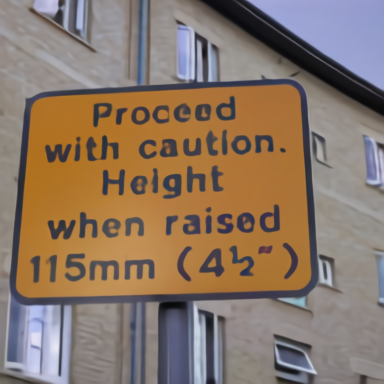} &
                \includegraphics[width=0.19\linewidth]{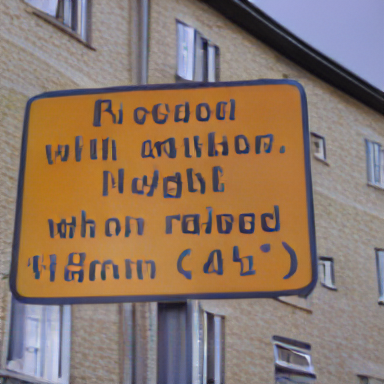} &
                \includegraphics[width=0.19\linewidth]{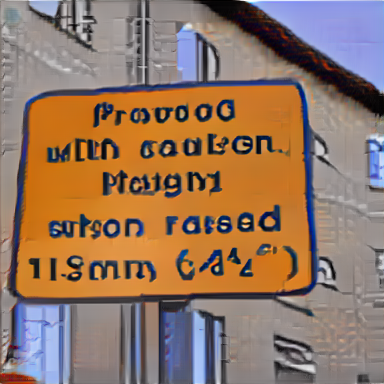} &
                \includegraphics[width=0.19\linewidth]{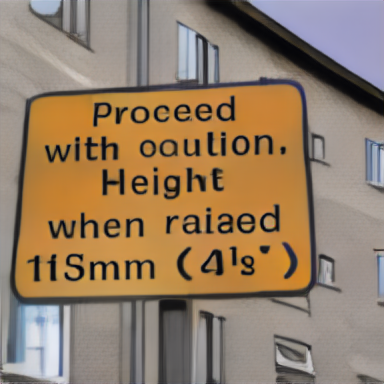}
                \\
                
                \includegraphics[width=0.19\linewidth]{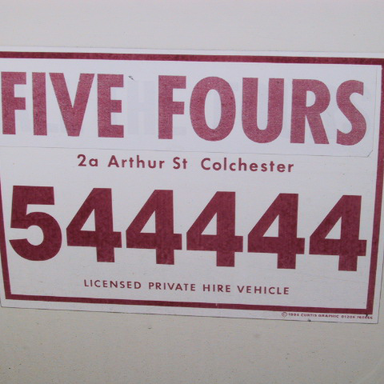} &
                \includegraphics[width=0.19\linewidth]{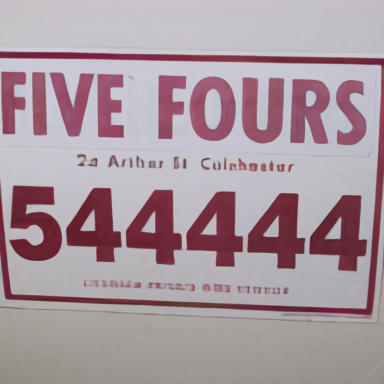} &
                \includegraphics[width=0.19\linewidth]{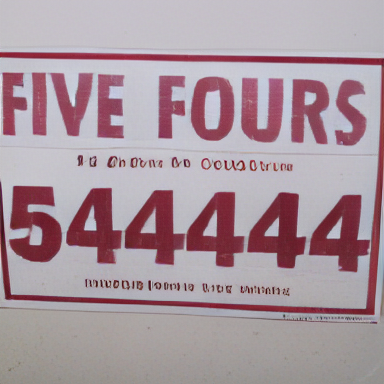} &
                \includegraphics[width=0.19\linewidth]{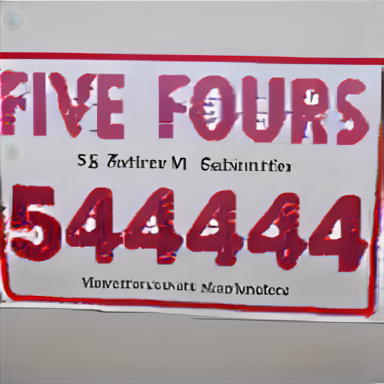} &
                \includegraphics[width=0.19\linewidth]{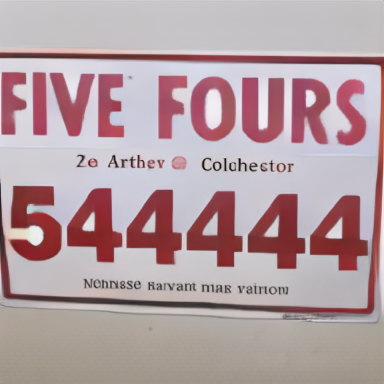}
                \\
                
                \bottomrule
            \end{tabular}
        }
    
    \end{center}
    \caption{Reconstructed images from ICDAR13 test set.}
    \label{table:appendix_ICDAR2013_table2}
\end{table*}
\vfill\null

\section{OCR Perceptual loss in PyTorch }
We present a Python implementation of the OCR Perceptual loss. The OCR perceptual loss operation accepts input and reconstructed images and has access to the OCR model that computes OCR features.

\begin{lstlisting}[language=Python, caption={Implementation of OCR Perceptual loss in Python (PyTorch)}]
import torch

def normalize_tensor(x,eps=1e-10):
    norm_factor=torch.sqrt(torch.sum(x**2,dim=1))
    return x/(norm_factor+eps)
    
def ocr_perceptual_loss(image, reconstruction):
    input_ocr_layers=ocr_model(image)
    rec_ocr_layers=ocr_model(reconstruction)
    
    ocr_loss=0
    for l in layers:
        in_feat=normalize(input_ocr_layers[l])
        rec_feat=normalize(rec_ocr_layers[l])
        
        diffs=(in_feat - rec_feat)**2
        diffs=diffs.sum(dim = 1)
        ocr_loss+=diffs.mean([2, 3])
    
    return ocr_loss
\end{lstlisting}

\end{appendices}

\end{document}